\documentclass{article}

\PassOptionsToPackage{numbers, compress}{natbib}

\usepackage[preprint]{neurips_2026}
\usepackage[utf8]{inputenc}
\usepackage[T1]{fontenc}
\usepackage{hyperref}
\usepackage{url}
\usepackage{booktabs}
\usepackage{amsfonts}
\usepackage{amsmath}
\usepackage{amssymb}
\usepackage{nicefrac}
\usepackage{microtype}
\usepackage{xcolor}
\usepackage[normalem]{ulem}
\usepackage{graphicx}
\usepackage{subcaption}
\usepackage{placeins}
\usepackage{tikz}
\usetikzlibrary{positioning,arrows.meta,backgrounds,calc,fit}
\graphicspath{{results/}}

\newcommand{\refsrc}{STCC}

\title{ALLMTS: Automated LLM-based Medical Triage System}

\author{%
Michał Remigiusz Janiszewski\textsuperscript{1}\thanks{Corresponding author}\phantom{\textsuperscript{*}}, 
Artur Stopa\textsuperscript{1}, 
Łukasz Lenkiewicz\textsuperscript{1,2},\\
\textbf{Wojciech Achtelik\textsuperscript{1,3},
Bartosz Gonczarek\textsuperscript{1,4}}\\[0.5em]
\textsuperscript{1}Vstorm, R\&D, Poland \\
\textsuperscript{2}Wroc\l{}aw University of Science and Technology, Wroc\l{}aw, Poland \\
\textsuperscript{3}Faculty of Computer Science, AGH University of Krak\'ow, Krak\'ow, Poland \\
\textsuperscript{4}University of Technology and Arts in Applied Sciences in Warsaw, Wrocław, Poland
}

\begin{document}

\maketitle

\begin{abstract}
Telephone triage is a safety critical task: from a brief intake note, a nurse must decide whether a caller needs an ambulance, an emergency department, an appointment, or self care at home. Queried directly, large language models (LLMs) are unreliable: they over-escalate routine cases and cannot justify their recommendations against an auditable clinical standard. We present the Automated LLM-based Medical Triage System (ALLMTS), which grounds every disposition in the same commercially maintained telehealth guidelines nurses use. 
Rather than asking an LLM to triage, ALLMTS makes it \emph{execute the guideline}: it selects the appropriate guideline and evaluates its triage assessment questions. The disposition follows deterministically from the first positive answer, using the same traversal rule a nurse follows.
We evaluate ALLMTS on 566 synthetic, expert-validated scenarios spanning 14 dispositions and designed to test specific guideline questions. ALLMTS achieves 0.95 accuracy, compared with 0.47 for a raw-LLM baseline.
On a curated 50-scenario subset released as an open benchmark, ALLMTS reaches 0.98 accuracy, on par with the 0.944 correct disposition rate that a five nurse expert panel recorded on those same scenarios. Critically, on that subset the baselines have over-triage rates of up to 0.50 (0.44 on the full dataset), whereas ALLMTS reduces the over-triage rate to at most 0.02 while maintaining low under-triage. These results support coupling LLM language understanding with executable clinical guidelines for triage from intake notes.
\end{abstract}

\section{Introduction}
\label{sec:intro}
Telephone and telehealth triage services are the first point of contact between patients and the healthcare system \cite{lake2017quality}. A triage nurse listens to a caller's complaint, selects the appropriate clinical guideline from a standardised protocol set, and works through the guideline's assessment questions, each tied to a level of care and ordered from immediate emergency care down to self care at home, until a positive finding at the most acute level fixes the recommended disposition. Telephone triage guideline sets vary in the number and naming of their disposition levels. The stakes of this decision are asymmetric: \emph{under-triage} sends a genuinely sick patient home \cite{giesen2007safety}, while \emph{over-triage} floods already crowded emergency departments with cases that could have waited for an office visit \cite{morley2018emergency}. Performing this task well requires clinical judgment, familiarity with a large guideline corpus, and consistency under high call volumes, conditions under which even experienced staff are subject to fatigue and inter-rater variability \cite{huibers2011safety,wheeler2015safety}.

Large language models (LLMs) offer an appealing way to automate this process: they encode broad medical knowledge \cite{singhal2023large,achiam2023gpt} and are already used across clinical decision support, diagnostics, and documentation \cite{thirunavukarasu2023large,kim2024health}, including emergency department triage \cite{triageagent,gaber2025evaluating}. However, LLMs remain unreliable as autonomous clinical decision makers \cite{hager2024evaluation}. Their parametric knowledge, however broad, may not align with the specific standards that a triage service is required to follow, while their free form reasoning provides no auditable link between those standards and the resulting recommendation.

In telephone triage the clinical expertise is already codified: commercially maintained protocol sets \cite{thompson2026adult,reder2026briggs,wheeler2009telephone} specify, for each presenting complaint, which questions to ask, in what order and a mapping from answer to disposition. We focus on faithful \emph{execution} of these guidelines once the intake note has been written: selecting an applicable guideline and assessing its triage questions against the information in the note.

We propose ALLMTS, an automated LLM-based medical triage system which decomposes triage into narrow, schema constrained language tasks: it extracts patient demographics and reasons for visit from the free text scenario, narrows the guideline corpus by metadata filtering and dense retrieval, selects a guideline under a runtime constrained output space that makes off corpus answers impossible and evaluates every triage assessment question (TAQ) of every disposition level concurrently. TAQ answers are binary under a closed world assumption, with boolean semantics for compound questions. The disposition is not generated by the model at all: it follows deterministically from the first positive TAQ, the same traversal rule a nurse applies, which makes every recommendation traceable to a specific guideline question and answer.

We evaluate ALLMTS on 566 clinical scenarios authored by the clinical content team at LaGrange Medical Software (Schmitt-Thompson Clinical Content) and validated by an expert triage nurse panel, covering 14 disposition labels across the full acuity spectrum. Against raw-LLM and web-search baselines and across four label granularities, ALLMTS reaches 0.95 disposition accuracy, compared with 0.47 for the raw LLM, cutting the over triage rate from 0.44 to 0.02. On a curated 50-scenario subset, which we release as an open benchmark, it reaches 0.98. A repeated-run analysis further shows the baseline's errors are systematic rather than sampling noise.

In summary, our contributions are:
\begin{itemize}
  \item \textbf{A guideline executing triage architecture.} We operationalise a complete commercial telehealth guideline corpus as an LLM pipeline combining retrieval based guideline selection, runtime schema locked TAQ (Triage Assessment Question) evaluation, deterministic first-positive-TAQ disposition, and guideline-to-guideline SMAG (See More Appropriate Guideline) re-routing, in which the LLM performs only language understanding and every clinical decision remains auditable.
 \item \textbf{An expert-validated evaluation and open benchmark.} We evaluate on 566 expert-validated scenarios spanning 14 dispositions, and together with LaGrange Medical Software (Schmitt-Thompson Clinical Content), publicly release the curated 50-scenario subset with reference dispositions as \emph{STCC Adult After-Hours Telehealth Triage 2026 -- 50 Clinical Scenarios}~\cite{allmts_triage_dataset}, an open benchmark for automated telehealth triage.
  \item \textbf{A clinically grounded comparison.} We benchmark against raw-LLM and web-search baselines, alongside a guideline-context ablation across four label granularities, reporting accuracy, weighted F1, the clinically decisive under-triage and over-triage rates as well as run-to-run stability.
\end{itemize}

\section{Related Works}

Large language models (LLMs) are increasingly adopted in healthcare for report generation, pharmaceutical assistance, clinical decision support, diagnostics, and education \cite{kim2024health,jin2024agentmd,thirunavukarasu2023large,van2024adapted,liu2023pharmacygpt,karttunen2023large,almubark2025exploring,yu2024simulated}. Many of these applications build on general purpose model families such as GPT, Claude, and Gemini \cite{achiam2023gpt,anthropic2024claude,team2023gemini}. Despite broad reasoning capabilities, they remain unreliable in specialized medical settings that demand domain expertise and adherence to task specific constraints \cite{singhal2023large,hager2024evaluation}. A central challenge is keeping them aligned with evolving clinical evidence: fine-tuning requires large, high-quality datasets that are costly to curate and constrained by privacy, while in context learning \cite{dong2024survey} only guides the use of information already encoded by the model without supplying external evidence. Retrieval-Augmented Generation (RAG) addresses this by augmenting the context with retrieved documents \cite{kechun2026retrieval,gargari2025enhancing}. Medical frameworks such as MEDIQ \cite{li2024mediq} and UOT \cite{hu2024uncertainty} add adaptive questioning and uncertainty-aware retrieval, and agentic systems let models decide when to retrieve or invoke tools \cite{jin2024genegpt}, with recent surveys giving a broader view \cite{amugongo2025retrieval, liang2025reasoning, plaat2025agentic}. In the distinct setting of long form evidence synthesis, MedProbeBench finds deep research systems limited in evidence integration and claim level citation support \cite{liu2026medprobebench}.

Outside the LLM literature, commercial virtual triage engines built on curated knowledge bases are already deployed alongside live nurse triage. A vignette based comparison of one against rules based telephone protocols reported comparable care referral accuracy and identical safety performance \cite{gellert2024comparative}. These systems interview the patient directly through their own knowledge base rather than executing a published guideline against an existing intake note, which is the task we address.

Recent work has explored grounding LLM-based clinical decision support in explicit guidelines. CPGPrompt converts narrative or diagrammatic guidelines into auditable decision trees for LLM traversal and evaluates the approach on guidelines for headache, lower back pain, and prostate cancer \cite{deng2026cpgprompt}. MedGUIDE evaluates adherence to 55 National Comprehensive Cancer Network (NCCN) decision trees spanning 17 cancer types \cite{li2025medguide}. Whereas CPGPrompt is evaluated on three disease specific guidelines and MedGUIDE is restricted to oncology, ALLMTS operationalises a maintained corpus of 314 adult telehealth triage guidelines spanning a broad range of presenting complaints.

Closer to our setting, several methods apply LLMs directly to triage decision-making. TriageAgent \cite{triageagent} assigns Emergency Severity Index levels through iterative multi-agent debate with confidence weighted consensus. A KTAS-based decision support system \cite{han2024ktascdss} role-plays a nurse, physician, pharmacist, and attending as four agents sharing one backbone LLM. C-PATH \cite{shi2025cpath} fine-tunes LLaMA-3 on synthetic dialogue data to route patients to a department and CLARITY \cite{shaposhnikov2025clarity} wraps a finite state dialogue manager around small supervised classifiers and one generative LLM. Despite their different mechanisms, all four ultimately rely on the LLM's intrinsic parametric knowledge to produce or refine the disposition itself. ALLMTS instead derives it from the guidelines, grounding recommendations in externally validated content and making each traceable to a particular \textit{triage assessment question} rather than to an ensemble vote or model internals.

\section{Method}
\label{sec:method}
\newcommand{\guidelinefootnote}{\footnote{The triage system described in this paper was
developed using the STCC Adult After-Hours Telehealth Triage Guidelines available from
Schmitt-Thompson Clinical Content (STCC) and owned by LaGrange Medical
Software, Inc.\ (LMS). The authors' organisation performed this work in
its capacity as a software development partner under a nondisclosure
agreement with LMS. The guidelines are used with permission and remain the
exclusive intellectual property of LMS. No proprietary clinical content
from the guidelines is reproduced or disclosed in this publication.}}

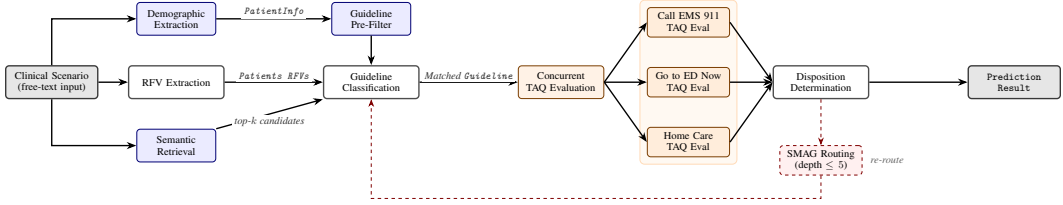
\begin{figure}[tbp]
\centering
\resizebox{\linewidth}{!}{%
\begin{tikzpicture}[
  node distance = 0.9cm and 0.9cm,
  mainbox/.style = {
    rectangle, rounded corners = 3pt, draw = black!70, very thick,
    fill = white, text centered, font = \small,
    minimum height = 1.0cm, text width = 2.7cm, inner sep = 3pt
  },
  endbox/.style = {
    rectangle, rounded corners = 3pt, draw = black!78, very thick,
    fill = black!10, text centered, font = \small,
    minimum height = 1.0cm, text width = 2.6cm, inner sep = 3pt
  },
  prebox/.style = {
    rectangle, rounded corners = 3pt, draw = blue!45!black, thick,
    fill = blue!8, text centered, font = \small,
    minimum height = 1.0cm, text width = 2.2cm, inner sep = 3pt
  },
  taqbox/.style = {
    rectangle, rounded corners = 3pt, draw = orange!48!black, thick,
    fill = orange!14, text centered, font = \small,
    minimum height = 0.9cm, text width = 2.3cm, inner sep = 3pt
  },
  gbox/.style = {
    rectangle, rounded corners = 3pt, draw = orange!48!black, thick,
    fill = orange!10, text centered, font = \small,
    minimum height = 1.0cm, text width = 2.4cm, inner sep = 3pt
  },
  smagbox/.style = {
    rectangle, rounded corners = 3pt, draw = red!50!black, thick, dashed,
    fill = red!6, text centered, font = \small,
    minimum height = 0.9cm, text width = 2.3cm, inner sep = 3pt
  },
  arrow/.style    = {->, >=Stealth, very thick},
  smagarrow/.style = {->, >=Stealth, thick, draw = red!50!black, dashed},
  annot/.style    = {font = \small\itshape, text = gray!80!black},
  datalbl/.style  = {font = \small\itshape, text = black!80,
                     fill = white, inner sep = 2pt},
]

\node[endbox]                          (input)  {Clinical Scenario\\(free-text input)};
\node[mainbox, right=of input]         (rfv)    {RFV Extraction};
\node[mainbox, right=3.0cm of rfv]     (cls)    {Guideline Classification};
\node[gbox,    right=3.0cm of cls]     (gather) {Concurrent TAQ Evaluation};
\node[taqbox,  right=1.3cm of gather]  (taq2)   {Go to ED Now\\TAQ Eval};
\node[taqbox,  above=of taq2]          (taq1)   {Call EMS 911\\TAQ Eval};
\node[taqbox,  below=of taq2]          (taq3)   {Home Care\\TAQ Eval};
\node[mainbox, right=1.3cm of taq2]    (disp)   {Disposition Determination};
\node[endbox, right=3.0cm of disp]     (output) {\texttt{Prediction}\\\texttt{Result}};

\node[prebox, above=0.9cm of rfv]      (demog)  {Demographic\\Extraction};
\node[prebox, above=0.9cm of cls]      (filt)   {Guideline\\Pre-Filter};
\node[prebox, below=0.9cm of rfv]      (retr)   {Semantic\\Retrieval};

\node[smagbox, below=1.4cm of disp]    (smag)   {SMAG Routing\\(depth $\leq 5$)};

\draw[arrow] (input) -- (rfv);
\draw[arrow] (rfv)   -- node[datalbl,above]{\texttt{Patients RFVs}} (cls);
\draw[arrow] (cls)   -- node[datalbl,above]{Matched \texttt{Guideline}} (gather);

\draw[arrow] (gather.east) -- (taq1.west);
\draw[arrow] (gather.east) -- (taq2.west);
\draw[arrow] (gather.east) -- (taq3.west);

\draw[arrow] (taq1.east) -- (disp.west);
\draw[arrow] (taq2.east) -- (disp.west);
\draw[arrow] (taq3.east) -- (disp.west);

\draw[arrow] (disp) -- (output);

\draw[arrow] (input) |- (demog.west);
\draw[arrow] (demog) -- node[datalbl,above]{\texttt{PatientInfo}} (filt);
\draw[arrow] (filt)  -- (cls);

\draw[arrow] (input) |- (retr.west);
\draw[arrow] (retr.north east) -- node[datalbl,pos=0.5,below]{top-$k$ candidates} (cls.south west);

\draw[smagarrow] (disp.south) -- (smag.north);
\draw[smagarrow] (smag.south) -- ++(0,-0.7) -| (cls.south);

\node[annot, right = 0.12cm of smag] {re-route};

\begin{scope}[on background layer]
  \node[draw = orange!28, fill = orange!5, rounded corners = 4pt,
        fit = (taq1)(taq2)(taq3), inner sep = 6pt] (grptaq) {};
\end{scope}

\end{tikzpicture}%
}
\caption{%
  Architecture of the ALLMTS pipeline.
  Grey nodes are input/output, white nodes are the core sequential stages, blue nodes are the pre-classification stages that feed classification, amber nodes form the concurrent TAQ evaluation block, the dashed red box marks conditional SMAG re-routing. Intermediate data artifacts are shown as labels on the arrows rather than as separate boxes. The TAQ evaluation stage runs all disposition levels concurrently.
}
\label{fig:scp-pipeline}
\end{figure}

In ALLMTS, the LLM executes a clinical guideline\guidelinefootnote. The model extracts structured fields from the intake note and answers one guideline question at a time against that note. The disposition is determined by a fixed rule, and each recommendation can be traced to the corresponding guideline question and answer.
Figure~\ref{fig:scp-pipeline} presents the pipeline.

\subsection{Task setup}
\label{sec:task-setup}

The input is a free text intake note $x$: the record a triage nurse writes during a
call, laid out as chief complaint, history, review of systems, and past medical
history. $\mathcal{D}$ is the list of dispositions available within the selected
guideline. The output is a disposition $\hat{d} \in \mathcal{D}$ together with the TAQ
that settled the case.
The model receives $x$ as written, without prior parsing, and extracts demographics, reasons for visit, and answers the guideline's triage assessment questions based on the note.

The input format, with a worked example
(Figure~\ref{fig:example-scenario}), is given in Appendix~\ref{app:input-format}.

\subsection{Pre-classification processing}
\label{sec:preclass}

Two extraction steps run before guideline classification. The first recovers the
patient's age and biological sex, taking only explicitly stated values (no inference) and leaving either field null when unstated. If neither is recoverable, pre-filtering is
bypassed and all guidelines are retained. The second extracts the reason for visit
(RFV), used both for classification and as context for TAQ evaluation: a \emph{primary}
RFV plus a list of \emph{secondary} ones, each a free text reason with a type
(\emph{Exposure}, \emph{Injury}, \emph{Pain}, or \emph{Symptom}) and an optional onset.
The primary RFV is the most clinically urgent presenting complaint. An explicit chief complaint is adopted as primary unless a clearly more urgent finding is present.

\subsection{Guideline selection}\label{sec:guideline-selection}

Selection maps the scenario and its RFV to one guideline in two steps. First, standard
embedding-based dense retrieval \cite{karpukhin2020dense,lewis2020retrieval} reduces the
corpus: the scenario is augmented with the primary RFV and the $k$ guidelines with the
highest cosine similarity form the candidate set $\mathcal{C}$. This step is necessary
because the system uses 314 guidelines, too many to pass into context. We provide configuration details in
Appendix~\ref{app:retriever}.

Second, the selection model reads each candidate's title, definition, category, gender,
type, system, and anatomy alongside the scenario, the RFV, and any structured patient
information, and returns a single title from $\mathcal{C}$. It follows the priority the
guideline authors specify: the primary RFV must fall within the guideline's
chief complaint scope, the definition must fit the presentation, and category and gender
must match the patient. When two guidelines apply, the more specific anatomical one wins
and a life threatening condition takes precedence. The model returns none only when
nothing in the corpus is clinically relevant.

\subsection{Triage assessment question evaluation}

TAQ evaluation is the clinical reasoning core of the pipeline. A guideline $g$ holds an ordered list of dispositions $D = \langle d_1, \ldots, d_m \rangle$ from highest to lowest acuity (Call EMS 911 Now, Go to ED Now, \ldots, Home Care), and each disposition $d_j$ carries its own ordered list of TAQs, $Q_j = \langle q_{j,1}, \ldots, q_{j,n_j} \rangle$. The model answers each TAQ against the scenario, deciding whether the condition it describes is present. The pipeline evaluates every disposition at once, producing one answered checklist per level.

Each disposition gets its own answer schema, with one field per TAQ. The question text, its exceptions, and its sub-clauses are all fixed in that schema, so the model can only answer each question as written: it receives the full checklist, pre-populated with default answers, and fills it in.

Each answer $a_i \in \{\texttt{true}, \texttt{false}\}$, and anything the scenario does
not mention counts as false (a closed-world assumption), which suits the structured
intake scenarios evaluated here. A TAQ may hold several sub-clauses joined by
conjunction or disjunction, encoded in the database with a bracket notation. The pipeline
splits these and folds them into one answer with standard boolean logic. Some TAQs also
carry an exception clause, parsed when the guideline loads, which forces the answer to
\texttt{false} regardless of the primary answer. One TAQ illustrating both:

\texttt{[1] Vomiting AND [2] contains red blood or black ("coffee ground") material
(Exception: Few red streaks in vomit that only happened once.)}

\subsection{Disposition determination and SMAG routing}\label{sec:disposition}

The pipeline merges the checklists in order of descending acuity and takes the first
positive TAQ as the recommended disposition: the highest acuity disposition with any
positive answer, and within it the first positive TAQ is the one that settled the case.
If no TAQ is positive anywhere, the pipeline recommends none.

Individual TAQs may also carry a link to an alternate guideline (SMAG routing). If the
first positive TAQ carries such a link, the pipeline switches to the linked algorithm and
re-evaluates its TAQs, up to a depth limit (default 5) with a visited set preventing
cycles, recording the chain of switched guidelines and each triggering TAQ.

The pipeline returns a single structured prediction carrying the disposition, the
selected guideline, the full answered TAQ checklist, and the retrieval and SMAG
metadata needed to audit the decision. Appendix~\ref{app:output-schema} lists the
fields.

\section{Experimental setup}\label{sec:experimental-setup}

\subsection{Models and inference settings}\label{sec:models}

All methods use \texttt{gpt-5.2} through the Azure OpenAI Responses API with \emph{low} reasoning effort.

\subsection{Disposition label schema}\label{sec:label-schema}

All methods predict over the fine-grained disposition space $\mathcal{D}$, the full taxonomy used by the clinical guideline database. For grouped evaluation, we also project each disposition into three coarser views defined in the source disposition table by fixed lookups: \emph{Acuity} (5 clinical acuity bands), \emph{Urgency} (10 time to care ratings) and \emph{Urgency Group} (7 tiers collapsing those ratings). The 14 dispositions our dataset exercises reach all 5 acuity bands, 8 of the urgency ratings and 6 of the urgency-group tiers. The views were defined separately, so they do not nest into a single hierarchy: \emph{Call PCP Now}, for instance, is only \emph{Urgent} on acuity yet shares the \emph{High} urgency group with the \emph{Emergent} dispositions. Reporting all four therefore avoids committing the evaluation to one guideline's granularity. Table~\ref{tab:label-views} in Appendix~\ref{app:label-views} gives the mapping for every disposition the evaluation exercises.

\subsection{Dataset}\label{sec:dataset}

The evaluation dataset holds $566$ clinical scenarios, each a free text intake note in the form given in Section~\ref{sec:task-setup}. Every scenario carries a reference recommended disposition (its ground truth label), together with the reference guideline, primary reason for visit, and first positive TAQ used for auxiliary analysis. The scenarios are concise, matching real telephone triage intake: the median runs to $81$ words (Figure~\ref{fig:dist-length}).

The reference dispositions cover all $14$ classes of
Table~\ref{tab:label-views} and are markedly imbalanced
(Appendix~\ref{app:label-distribution}, Figure~\ref{fig:dataset-dist}). Five dispositions carry most of the volume ($12$--$15\%$ each), while a tail of six clinically valid but rare ones covers at most seven scenarios each, two of them occurring exactly once. This skew is why we weight F1 by support throughout the results (Section~\ref{sec:results}). Coarsening concentrates the mass into better populated bands but keeps the head-heavy shape. Together with LMS (STCC), we publicly release the curated 50-scenario subset, with each scenario's reference disposition, as \emph{STCC Adult After-Hours Telehealth Triage 2026 -- 50 Clinical Scenarios}~\cite{allmts_triage_dataset}, an open benchmark for automated telehealth triage.

\subsubsection{Clinical scenario development}\label{sec:scenario-development}

The clinical scenarios were written and validated by the clinical content team at LaGrange Medical Software (Schmitt-Thompson Clinical Content), the authors of the underlying triage guidelines, through a structured 14-step process. Each scenario is a simulated patient case built to test one triage assessment question within a single adult telehealth guideline, and carries enough clinical detail to fix a primary reason for visit, the first positive TAQ, and the correct reference disposition. Scenarios follow a standard chief-complaint / history / review-of-systems / past-medical-history format, with set conventions for age, sex distribution, symptom onset, length, and reading level. Chief complaints use varied, patient-centered language rather than guideline terminology, and omit names, occupations, and cultural or socioeconomic identifiers.

A large language model produced the initial drafts under structured prompts. An experienced Nurse Editor then reviewed and refined each one, keeping full editorial control over clinical accuracy and guideline alignment. A second Nurse Editor reviewed each scenario independently, and scenarios were revised or discarded when the two disagreed. The survivors went to a panel of five expert telehealth triage nurses for validation. On the 50-scenario subset released here, the nurses achieved a correct disposition rate of $0.944$. We use this rate as a human reference point in Section~\ref{sec:human-benchmark}. Validated scenarios were approved by the Senior Medical Editor and held in a version controlled content management system, from which the benchmark set was drawn.

\subsection{Baseline methods}\label{sec:baselines}

We compare the ALLMTS pipeline against two baselines. They both emit the same output as ALLMTS, a disposition $\hat{d} \in \mathcal{D}$ plus a one-sentence reasoning string kept for audit only, and are scored the same way.

\subsubsection{Raw LLM}
A single structured call with no external context. The model reads the scenario as the user prompt, with a system prompt listing the admissible disposition labels, their urgency definitions, and a clinical glossary, and imposing a closed world assumption: any finding not explicitly stated is treated as absent. No guideline knowledge or retrieval is injected, so the model relies entirely on its pre-training.

\subsubsection{Web search}
The model is given a built-in \texttt{web\_search} tool (via the Azure OpenAI Responses API) and instructed to issue at least one query for clinical triage guidelines matching the chief complaint before committing to a disposition, then conditions on the retrieved sources; queries and source URLs are pulled from the response trace for reproducibility. This tests whether generic web evidence, rather than the specific maintained corpus, is enough to close the gap.

\subsection{Guideline-context ablation}\label{sec:guideline-context}

This keeps ALLMTS's guideline selection step (Section~\ref{sec:guideline-selection}) but removes explicit TAQ evaluation and SMAG re-routing. The full text of the selected guideline, its severity scales, initial assessment questions, and every TAQ grouped under its disposition heading, is pasted into the prompt verbatim, and the model picks one disposition from that guideline's own dispositions in a single call. It does not require explicit per-TAQ answers to determine the disposition using the first-positive rule of Section~\ref{sec:disposition}, isolating what these execution stages add over simply giving the model the selected guideline to read. Both methods are evaluated on the final disposition. ALLMTS additionally produces explicit per-TAQ answers that determine its disposition through the first-positive rule; the ablation predicts the disposition directly, without requiring these intermediate answers.

\section{Results}
\label{sec:results}

We report on two evaluation sets, running every method on both: a curated set of $n=50$
held out scenarios, and the full $n=566$ dataset, a strict superset of it. The larger set
also exercises a wider vocabulary: 14 reference dispositions versus 10 in the subset
(Table~\ref{tab:label-views}).

Each method is scored on accuracy, weighted F1, and three triage error rates:
\emph{under-triage} (a less urgent label than the reference, also including no-disposition outcomes),
\emph{over-triage}
(a more urgent label), and \emph{discordance} (any mismatch). F1 is weighted by support
rather than macro averaged because six of the 14 classes carry at most seven scenarios
(Section~\ref{sec:dataset}), and a macro score would let that tail dominate the head
classes. In \emph{ad-hoc} mode the model predicts directly in a view's label space. In
\emph{post-hoc} mode the disposition prediction is projected into a coarser view. We
report each (method, view) cell split by mode: 50-scenario results in
Table~\ref{tab:results-50}, full dataset results in Table~\ref{tab:results-566}.

ALLMTS can additionally return a \emph{No positive TAQ found} outcome, when TAQ
evaluation reaches the end of the matched guideline with no positive answer for the
first-positive-TAQ rule (Section~\ref{sec:disposition}) to fire on. The baselines cannot,
as they always emit some label from $\mathcal{D}$. We keep such cases verbatim rather
than forcing a fallback: they count as discordant under every view and form a trailing
class in the ALLMTS confusion matrices, occurring in 9 of 566 cases ($1.6\%$). For scoring, these outcomes also count as under-triage,
not over-triage; this convention does not assign them a clinical
urgency level.

\begin{table}[t]
\caption{50-scenario set: accuracy / weighted F1 (Acc/F1, higher is better) and
under-/over-triage rates (UT/OT, lower is better). In the \emph{ad-hoc} panel the model
predicts directly in each view's label space. In the \emph{post-hoc} panel each
coarse-view label is projected from the fine-grained disposition prediction. Only the
baselines and the ablation populate the ad-hoc coarse views, since ad-hoc ALLMTS
predicts on the disposition space directly. Discordance is the sum of the two
triage-error rates.}
\label{tab:results-50}
\centering
\footnotesize
\setlength{\tabcolsep}{3pt}
\begin{tabular}{lcc|cc|cc|cc}
\toprule
& \multicolumn{2}{c|}{ALLMTS} & \multicolumn{2}{c|}{Guideline Ctx} & \multicolumn{2}{c|}{Raw LLM} & \multicolumn{2}{c}{Web Search}\\
View & Acc/F1 & UT/OT & Acc/F1 & UT/OT & Acc/F1 & UT/OT & Acc/F1 & UT/OT \\
\midrule
\multicolumn{9}{c}{\textbf{Ad-hoc}}\\
\midrule
Disposition   & .98/.98 & .02/.00 & .94/.94 & .00/.06 & .42/.40 & .14/.44 & .38/.35 & .14/.48 \\
Acuity        & --      & --      & .94/.94 & .02/.04 & .66/.66 & .04/.30 & .72/.72 & .06/.22 \\
Urgency       & --      & --      & .84/.81 & .04/.12 & .52/.50 & .14/.34 & .46/.43 & .16/.38 \\
Urgency Group & --      & --      & .76/.76 & .04/.20 & .58/.58 & .12/.30 & .58/.59 & .16/.26 \\
\midrule
\multicolumn{9}{c}{\textbf{Post-hoc}}\\
\midrule
Acuity        & .98/.98 & .02/.00 & .96/.96 & .00/.04 & .62/.61 & .04/.34 & .54/.52 & .06/.40 \\
Urgency       & .98/.98 & .02/.00 & .94/.94 & .00/.06 & .44/.42 & .10/.46 & .38/.35 & .12/.50 \\
Urgency Group & .98/.98 & .02/.00 & .94/.94 & .00/.06 & .54/.52 & .06/.40 & .46/.41 & .08/.46 \\
\bottomrule
\end{tabular}
\end{table}

\begin{table}[t]
\caption{Full 566-scenario dataset (superset of the 50-scenario set), scored and laid
out as in Table~\ref{tab:results-50}.}
\label{tab:results-566}
\centering
\footnotesize
\setlength{\tabcolsep}{3pt}
\begin{tabular}{lcc|cc|cc|cc}
\toprule
& \multicolumn{2}{c|}{ALLMTS} & \multicolumn{2}{c|}{Guideline Ctx} & \multicolumn{2}{c|}{Raw LLM} & \multicolumn{2}{c}{Web Search}\\
View & Acc/F1 & UT/OT & Acc/F1 & UT/OT & Acc/F1 & UT/OT & Acc/F1 & UT/OT \\
\midrule
\multicolumn{9}{c}{\textbf{Ad-hoc}}\\
\midrule
Disposition   & .95/.96 & .04/.02 & .90/.90 & .05/.04 & .47/.48 & .10/.44 & .50/.51 & .11/.39 \\
Acuity        & --      & --      & .94/.94 & .02/.04 & .67/.67 & .08/.25 & .70/.70 & .08/.22 \\
Urgency       & --      & --      & .89/.88 & .04/.07 & .56/.55 & .15/.29 & .54/.53 & .18/.28 \\
Urgency Group & --      & --      & .86/.86 & .06/.09 & .60/.61 & .16/.25 & .61/.62 & .16/.23 \\
\midrule
\multicolumn{9}{c}{\textbf{Post-hoc}}\\
\midrule
Acuity        & .96/.97 & .03/.02 & .94/.94 & .03/.03 & .61/.60 & .04/.35 & .63/.63 & .05/.31 \\
Urgency       & .95/.96 & .04/.02 & .91/.91 & .05/.04 & .49/.49 & .08/.43 & .53/.52 & .08/.39 \\
Urgency Group & .96/.97 & .03/.02 & .93/.93 & .04/.04 & .57/.56 & .05/.38 & .60/.59 & .06/.34 \\
\bottomrule
\end{tabular}
\end{table}

\subsection{ALLMTS versus baselines}

Our method leads on accuracy and weighted F1 in every cell of both tables. We take
the two evaluation sets in turn, then look at how the baselines fail.

\paragraph{50-scenario set.} ALLMTS reaches $0.98$ accuracy and F1 on the fine-grained
disposition space, against $0.42/0.40$ (Raw LLM) and $0.38/0.35$ (Web Search). It holds
$0.98$ across the coarser views, where the best baseline cell tops out at $0.72$ (Web
Search, ad-hoc acuity). Its only error is a single under-triage, with no over-triage.

\paragraph{566-scenario set.} The ranking is unchanged on the superset
(Table~\ref{tab:results-566}, Figure~\ref{fig:cm-566}). ALLMTS scores $0.95$ disposition
accuracy and up to $0.96$ across the projected views, against $0.47$ (Raw
LLM) and $0.50$ (Web Search) on disposition and a $0.70$ best view baseline ceiling. The
$11\times$-larger sample reduces ALLMTS accuracy by $0.03$, consistent with the full set's
harder long-tail dispositions and the $1.6\%$ of cases returning \emph{No positive TAQ
found} (an automatic discordance).

\paragraph{Baselines} Most baseline errors are over-triage: on full-set disposition, the Raw LLM over-triages $0.44$ of cases and Web Search $0.39$. Figures~\ref{fig:cm} and~\ref{fig:cm-566} in Appendix~\ref{app:confusion-matrices} show this as a spread above the diagonal, so the baselines usually err by sending the patient
somewhere more acute than the reference.

Web Search shows mixed results relative to the Raw LLM across views: on the full set it is
ahead on disposition ($0.50$ vs.\ $0.47$) and acuity ($0.70$ vs.\ $0.67$) but behind on
urgency, and at $n=50$ it is behind on both disposition and urgency. Either way it stays
at least $0.45$ below ALLMTS on disposition, so generic retrieval is not what closes the
gap. Baseline accuracy rises as fine distinctions collapse (Raw LLM: disposition $0.47
\rightarrow$ urgency group $0.60 \rightarrow$ acuity $0.67$), and both baselines do
better predicting a coarse label directly than projecting one from their disposition
prediction (Raw LLM urgency: $0.56$ ad-hoc vs.\ $0.49$ post-hoc), because projection
inherits the errors of a noisy disposition.

\paragraph{Guideline-context ablation.} Handing the model the selected guideline verbatim
(Section~\ref{sec:guideline-context}) recovers most, but not all, of the gap: $0.90$
ad-hoc disposition accuracy on the full set against ALLMTS's $0.95$ and the $0.47$ and
$0.50$ of the Raw LLM and Web Search, so the retrieved guideline, not parametric
knowledge or generic retrieval, carries the bulk of the improvement. Structured traversal
still increases accuracy by $0.02$ to $0.05$ in every like-for-like cell of
Table~\ref{tab:results-566} (disposition $0.90$ versus $0.95$, post-hoc acuity $0.94$
versus $0.96$, urgency $0.91$ versus $0.95$, urgency group $0.93$ versus $0.96$). The gap
is wider still where the ablation predicts a coarse label directly (ad-hoc urgency
$0.89$, urgency group $0.86$, against ALLMTS's projected $0.95$ and $0.96$): reading the
guideline holistically loses the ordering the first-positive rule enforces, and its
errors skew toward over-triage (ad-hoc urgency group $0.09$, against $0.02$ for ALLMTS
projected into the same view).

\paragraph{Human reference point.}\label{sec:human-benchmark} The validation process
behind the dataset (Section~\ref{sec:scenario-development}) supplies a human benchmark on
the 50-scenario subset: the five expert triage nurses achieved a correct disposition rate of $0.944$.

\subsection{Run-to-run stability}

A single run conflates systematic miscalibration with run-to-run noise, so we repeat all four methods on the acuity view ten times over the same 50 scenarios. Table~\ref{tab:stability} evaluates each method against the reference using accuracy and Cohen's $\kappa$, averaged over runs, and measures self agreement using Fleiss' $\kappa$, modal prediction consistency, and flip rate. Appendix~\ref{app:stability} visualises per scenario correctness across all ten runs for each method (Figures~\ref{fig:stability-scp}--\ref{fig:stability-ws}). We report $\kappa$ because the acuity distribution is skewed.

\begin{table}[t]
\caption{Run-to-run stability on the acuity view, 10 runs over the same 50 scenarios.
Acc is the per-run accuracy (mean $\pm$ s.d.). $\kappa_{\text{ref}}$ and
$\kappa_{\text{ref}}^{w}$ are unweighted and linearly weighted Cohen's $\kappa$ against
the reference label, averaged over runs. $\kappa_{\text{run}}$ is Fleiss' $\kappa$ across
the ten runs (self-agreement, independent of correctness). Cons.\ is mean
modal prediction consistency and Flip the fraction of scenarios whose prediction is not
identical on all ten runs. $10/10$ and $0/10$ count scenarios correct on every run and on
no run.}
\label{tab:stability}
\centering
\footnotesize
\setlength{\tabcolsep}{4pt}
\begin{tabular}{@{}lccccccc@{}}
\toprule
& \multicolumn{3}{c}{vs.\ reference} & \multicolumn{2}{c}{self-agreement} & \multicolumn{2}{c}{scenarios}\\
\cmidrule(lr){2-4}\cmidrule(lr){5-6}\cmidrule(lr){7-8}
Method & Acc & $\kappa_{\text{ref}}$ & $\kappa_{\text{ref}}^{w}$ & $\kappa_{\text{run}}$ / Cons. & Flip & $10/10$ & $0/10$ \\
\midrule
ALLMTS         & $.966 \pm .018$ & $.955$ & $.961$ & $.958$ / $.974$ & $.06$ & 47 & 0 \\
Guideline Ctx  & $.918 \pm .019$ & $.890$ & $.932$ & $.942$ / $.970$ & $.12$ & 43 & 1 \\
Raw LLM        & $.658 \pm .024$ & $.541$ & $.707$ & $.892$ / $.946$ & $.20$ & 28 & 12 \\
Web Search     & $.706 \pm .031$ & $.605$ & $.747$ & $.844$ / $.922$ & $.26$ & 28 & 9 \\
\bottomrule
\end{tabular}
\end{table}

ALLMTS is the most accurate and the most stable method, staying close to its single run
numbers in Table~\ref{tab:results-50}: $\kappa_{\text{ref}} = 0.955$ against $0.541$ for
the Raw LLM, a $0.41$ chance corrected gap wider than the $0.31$ accuracy gap suggests.
$47$ of $50$ scenarios are correct on every ALLMTS run and none is wrong on all ten,
against $28$ and $12$ for the Raw LLM. Web Search is the least self consistent
($\kappa_{\text{run}} = 0.844$), and the ablation sits between ALLMTS and the baselines
on every column. The residual error is systematic rather than sampling noise: self agreement is high for every
method ($\kappa_{\text{run}} \geq 0.84$) while reference agreement is not, so the
baselines converge on the same over-triaged label even when wrong, and weighting $\kappa$
by ordinal distance helps them far more than ALLMTS (Raw LLM $0.541 \rightarrow 0.707$,
ALLMTS $0.955 \rightarrow 0.961$).

\subsection{Comparison with prior methods}

We also benchmark against two prior LLM triage evaluations, Levine et al.~\cite{levine2023diagnostic} and Gaber et al.~\cite{gaber2025evaluating}, whose four-level and five-level (ESI) taxonomies differ from ours, by deterministically projecting each method's output and the \refsrc\ ground truth into a shared label space (Appendix~\ref{app:mappings}). Since both original studies used now-retired models, we re-ran their methods verbatim on the same backbone we use throughout (\texttt{gpt-5.2}), isolating the method's contribution from the model's.

Both prior methods (Table~\ref{tab:reproduction-metrics}) lean on intrinsic knowledge rather than external rules, biasing them toward over-triage and the middle of the scale. Their most common failure is escalating an urgent case to an emergent one. Levine's narrow four-level taxonomy masks the middle compression a more granular scale would expose, at accuracy comparable to Gaber's. Figure~\ref{fig:reproduction-bias} in Appendix~\ref{app:reproduction} shows the label distributions. ALLMTS exhibits neither bias.

\begin{table}[t]
\caption{Reproduced prior methods on the 50-scenario \refsrc\ set with \texttt{gpt-5.2}, scored in each method's native label space. Levine uses its four-level taxonomy while Gaber uses ESI.}
\label{tab:reproduction-metrics}
\centering
\small
\setlength{\tabcolsep}{6pt}
\begin{tabular}{@{}l c c c c c@{}}
\toprule
\textbf{Method} & \textbf{Acc} & \textbf{F1} & \textbf{Under} & \textbf{Over} & \textbf{Disc} \\
\midrule
Levine & 0.70 & 0.70 & 0.10 & 0.20 & 0.30 \\
Gaber & 0.58 & 0.55 & 0.18 & 0.24 & 0.42 \\
ALLMTS & 0.98 & 0.98 & 0.02 & 0.00 & 0.02 \\
\bottomrule
\end{tabular}
\end{table}

\subsection{Evaluation on related datasets}

We also ran ALLMTS directly on the in scope subsets of the two external vignette sets that best fit the adult teletriage setting, five runs each (Table~\ref{tab:allmts-external-vignettes} in Appendix~\ref{app:reproduction}). On Levine et al.~\cite{levine2023diagnostic} (43 of 48 in scope vignettes), it reaches only $0.41$ accuracy, with over half the errors being no answer rows. On Semigran et al.~\cite{semigran2015evaluation} (37 of 45 in scope) it reaches $0.58$, still far below its $0.95$--$0.98$ on our own dataset, and there most errors are confident wrong tier answers rather than no answers. In both cases, clinical experts deemed the vignettes not detailed enough: reaching a disposition within the clinical guidelines would require a clarifying question.

\section{Limitations}\label{sec:limitations}

ALLMTS is powered by the \refsrc\ clinical triage guidelines and was developed around them. We acknowledge the circularity of evaluating a system designed around some guidelines with a dataset built for the same guidelines, yet we chose to build a dedicated, high quality dataset for it, and we acknowledge the limited reproducibility that proprietary content imposes.

We excluded some available triage datasets because they encode information unavailable in a teletriage setting: the ESI Handbook derived vignettes used by TriageAgent \cite{triageagent} and the MIMIC-IV-Ext-CDS cases used by Gaber et al. \cite{gaber2025evaluating} embed bedside vitals a phone caller cannot report, and which the ground-truth acuity label was itself assigned with reference to. The small external vignette sets from Levine et al.\ \cite{levine2023diagnostic} and Semigran et al.\ \cite{semigran2015evaluation} were used for supplementary evaluation rather than as the primary benchmark (Table~\ref{tab:allmts-external-vignettes}). Pediatric cases were excluded, as the guidelines are adult only. Clinical experts also deemed several of these external vignettes too thin to judge without clarification questions.

The closed-world assumption ties the reported results to sufficiently detailed intake notes: an omitted finding and an explicitly negative one are not distinguished. An interactive variant of ALLMTS, with an explicit unknown state that triggers clarification questions to the caller or deferral to a triage nurse, is in development; its evaluation is left to future work.

Beyond the guideline-context ablation (Section~\ref{sec:guideline-context}), which removes TAQ traversal as a whole, we report no per stage ablation: the pipeline re-implements the procedure a triage nurse follows, and removing any single stage leaves the traversal undefined rather than degraded, so such an ablation would score a different system. The choices that can be varied while keeping traversal intact, namely the backbone model, the reasoning effort, and the retriever configuration, are reported in Section~\ref{sec:models} and Appendix~\ref{app:retriever}.

\section{Conclusions}

We presented ALLMTS, an automated medical triage system whose central design decision is to make the LLM execute an existing clinical standard rather than act as the clinical authority itself. Its pipeline decomposes telehealth triage into narrow, schema constrained language tasks (demographic and reason-for-visit extraction, retrieval based guideline selection, and per question TAQ evaluation), while the disposition follows deterministically from the first positive TAQ, so every recommendation is traceable to a specific guideline question and answer. We evaluated ALLMTS on 566 synthetic, expert-validated scenarios designed to test specific guideline questions and spanning 14 disposition labels. ALLMTS achieved $0.95$ disposition accuracy, compared with $0.47$ for the raw-LLM baseline. On the released 50-scenario subset, its accuracy was $0.98$, while a panel of five expert triage nurses achieved a correct disposition rate of $0.944$. Generic augmentation does not close that gap: web search helps only unevenly and is the least stable of the four methods, while an ablation that reads the selected guideline holistically instead of executing it recovers most of the gap but not all of it ($0.90$ against $0.95$). Access to the selected guideline accounts for most of the observed improvement, with explicit TAQ evaluation and SMAG re-routing together providing an additional gain on guideline-aligned scenarios. Performance was lower on the two external vignette sets, highlighting the need for further evaluation beyond guideline-aligned scenarios.

\section*{Acknowledgements}
The authors thank LaGrange Medical Software, Inc. (Schmitt-Thompson Clinical Content) for permitting the use of the STCC Adult After-Hours Telehealth Triage Guidelines and for authoring and clinically validating the evaluation scenarios used in this study.

\bibliographystyle{plainnat}
\bibliography{bibliography}

\appendix

\section{Label-view mapping}\label{app:label-views}

\begin{table}[!htbp]
\caption{Deterministic label view mappings. Each fine-grained disposition projects
independently into the acuity, urgency, and urgency-group label spaces used for
grouped evaluation. The table lists the 14 reference dispositions exercised by the
full 566-scenario dataset; the 50-scenario subset exercises 10 of them.
\emph{No positive TAQ found} is an ALLMTS-only outcome with no analogue in the fixed
baseline label space $\mathcal{D}$ and is kept verbatim rather than mapped.}
\label{tab:label-views}
\centering
\small
\begin{tabular}{llll}
\toprule
\textbf{Disposition ($\mathcal{D}$)} & \textbf{Acuity} & \textbf{Urgency} & \textbf{Urgency Group}\\
\midrule
Call EMS 911 Now                          & EMS 911                         & 911 Now           & Very High\\
Go to ED Now                              & Emergent                        & Emergent Now      & High\\
Go to L\&D Now                            & Emergent                        & Emergent Now      & High\\
Go to ED/UCC Now (or PCP Triage)          & Emergent                        & Within 1 Hour     & High\\
Go to L\&D Now (or PCP Triage)            & Emergent                        & Within 1 Hour     & High\\
Call PCP Now                              & Urgent                          & Within 1 Hour     & High\\
See HCP (or PCP Triage) Within 4 Hours    & Urgent                          & Within 4 Hours    & Moderate\\
See PCP Within 24 Hours                   & Urgent                          & Within 1 Day      & Moderate\\
Lab Test Only Within 24 Hours             & Urgent                          & Within 1 Day      & Moderate\\
Call PCP Within 24 Hours                  & Urgent                          & Within 1 Day      & Moderate\\
See PCP Within 3 Days                     & Non-Urgent or Routine           & Within 3 Days     & Low\\
Call PCP When Office is Open              & Non-Urgent or Routine           & Within 3 Days     & Low\\
See PCP Within 2 Weeks                    & Non-Urgent or Routine           & Within 2 Weeks    & Routine\\
Home Care                                 & Self Care or Health Information & Self Care at Home & Self Care\\
\midrule
\multicolumn{4}{l}{\footnotesize\emph{ALLMTS-only outcome (no analogue in $\mathcal{D}$):}}\\
\textit{No positive TAQ found} & \multicolumn{3}{l}{\footnotesize kept verbatim; see Section~\ref{sec:results}}\\
\bottomrule
\end{tabular}
\end{table}

\FloatBarrier
\section{Input format}\label{app:input-format}

\begin{figure}[!htbp]
\centering
\includegraphics[width=0.7\linewidth]{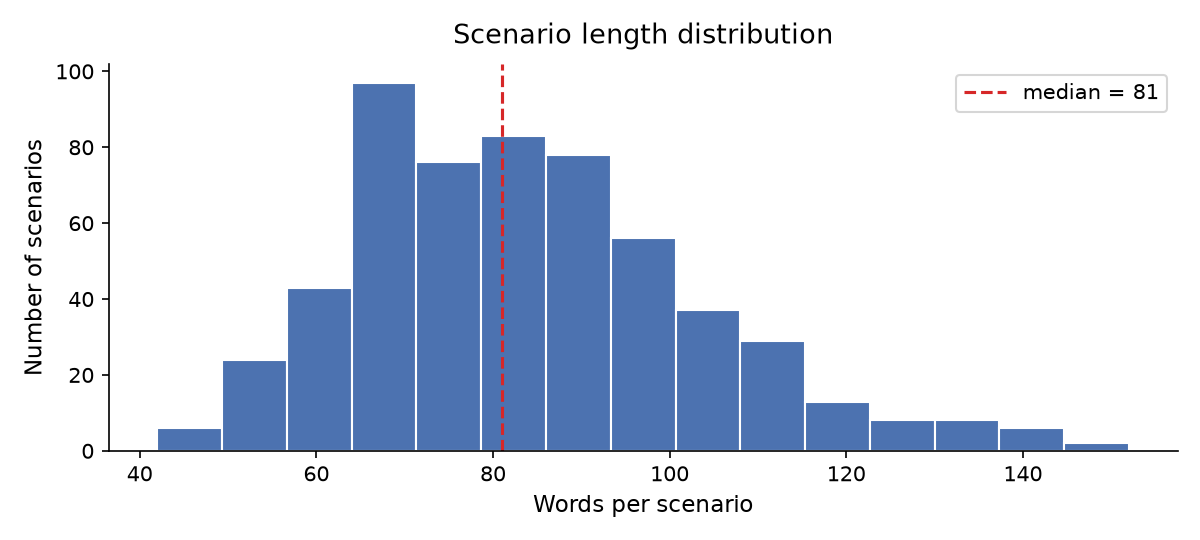}
\caption{Scenario-length distribution (words per scenario), with the median marked.}
\label{fig:dist-length}
\end{figure}

Scenarios arrive as plain text under four headings:

\begin{verbatim}
CHIEF COMPLAINT: <complaint>
HISTORY: <narrative>
REVIEW OF SYSTEMS: <systems>
PAST MEDICAL HISTORY: <history>
\end{verbatim}

The headings follow standard nursing intake documentation, but the
pipeline never parses them. They reach the model as part of the raw
prompt, which means a note can drop a section or reorder them without
breaking anything. When a chief complaint is present, the extractor treats
it as a strong prior on the primary reason for visit
(Section~\ref{sec:preclass}); no component requires it.
Figure~\ref{fig:dist-length} reports scenario lengths.
Figure~\ref{fig:example-scenario} gives a complete scenario from the released
50-scenario subset.

\begin{figure}[!htbp]
\centering
\fbox{\begin{minipage}{0.95\linewidth}
\footnotesize
\textbf{CHIEF COMPLAINT:} Brief arm weakness

\textbf{HISTORY:} A 59-year-old female had an episode of sudden-onset left arm weakness
30 minutes ago. She dropped a pan she was holding and could not lift her arm or move her
fingers. The weakness then completely resolved within five minutes. Her arm strength and
movement have returned to normal.

\textbf{REVIEW OF SYSTEMS:} Denies headache, speech problems, vision changes, back, or
arm pain.

\textbf{PAST MEDICAL HISTORY:} High cholesterol.
\end{minipage}}
\caption{An example scenario from the released 50-scenario subset, shown verbatim as the
pipeline receives it. Its reference disposition is \emph{Go to ED Now}.}
\label{fig:example-scenario}
\end{figure}

\FloatBarrier
\section{Output schema}\label{app:output-schema}

The pipeline returns a single structured prediction containing:

\begin{itemize}
  \item \textbf{disposition}: The recommended disposition linked to the First Positive Triage Assessment Question (FPTAQ).
  \item \textbf{guideline}: The selected clinical guideline.
  \item \textbf{taqs}: A \texttt{TriageAssessmentQuestionChecklist} with all evaluated
        TAQ answers across all dispositions.
  \item \textbf{rfv}: \texttt{PatientsReasonsForVisit} with primary and secondary reasons, types, and duration of onset.
  \item \textbf{guideline\_rationale}: LLM optional free-text justification for guideline selection.
  \item \textbf{retrieval\_result}:  Retrieval metadata (similarity scores, embedding model, index type, $k$).
  \item \textbf{smag\_switched\_from}: Ordered list of guideline titles traversed via SMAG routing.
  \item \textbf{smag\_first\_positive\_taq}: Ordered list of TAQ trigger questions for each SMAG switch.
\end{itemize}

\FloatBarrier
\section{Retriever configuration}\label{app:retriever}

The retriever is initialised once before inference. Each guideline $g_i$ in the
corpus $\mathcal{G}=\{g_1,\ldots,g_n\}$ is given a textual representation $s_i$,
embedded with $\phi$, then $L_2$-normalised and cached so it is not recomputed per
scenario. The implementation supports four representations:

\begin{itemize}
    \item \textbf{title}: the guideline title;
    \item \textbf{definition}: the title concatenated with the guideline definition;
    \item \textbf{context}: the title concatenated with selected metadata fields,
    such as anatomy and clinical context;
    \item \textbf{full}: the concatenation of all available guideline metadata.
\end{itemize}

All reported experiments use the title representation, embedded with an
Azure-hosted \texttt{text-embedding-3-large} ($d=3072$).

At inference, the scenario is augmented with the extracted primary reason for
visit,
\begin{equation*}
q = x + \text{`` Primary Reason for Visit: ''} + r_{\mathrm{primary}},
\end{equation*}
then embedded and $L_2$-normalised with the same model. Guidelines are ranked by
cosine similarity and the top-$k$ ($k=50$) passed to the selection model, with
query augmentation focusing retrieval on the dominant complaint.
Table~\ref{tab:retriever-hparams} summarises the configuration.

\begin{table}[!htbp]
\caption{Retriever configuration used in the experiments.}
\label{tab:retriever-hparams}
\centering
\begin{tabular}{ll}
\toprule
\textbf{Hyperparameter} & \textbf{Value} \\
\midrule
Embedding model & \texttt{text-embedding-3-large} \\
Embedding dimension & 3072 \\
Guideline representation & Title \\
Number of candidates & $k=50$ \\
Query augmentation & Primary reason for visit \\
Similarity function & Cosine similarity \\
Guideline embedding cache & Enabled \\
\bottomrule
\end{tabular}
\end{table}

Only the scenario-side embedding is computed per query; similarity scores and
retriever configuration are stored for auditing.

\FloatBarrier
\section{Output scale mappings}\label{app:mappings}

Table~\ref{tab:mappings} gives the deterministic projection used for the
head-to-head comparison in Section~\ref{sec:results}: each of the 10 reference
dispositions exercised by the 50-scenario set is routed through the acuity and urgency
views into Levine's four-level taxonomy and Gaber's ESI, so all methods can be scored in
a shared label space.

\begin{table}[!htbp]
\caption{Post-hoc projection of \refsrc\ ground-truth dispositions into the label spaces of the
reproduced methods. Rows are the 10 dispositions exercised by the 50-scenario set. Each disposition is routed through the \refsrc\ published \emph{acuity} (1--5)
and \emph{urgency} (10-label) views, then into Levine's four-level taxonomy and Gaber's ESI
(1--5). Levine is scored via the urgency path; Gaber via the \emph{via-acuity} path, a
near-identity alignment in which $\text{ESI}\,N=\text{acuity rating}\,N$.}
\label{tab:mappings}
\centering
\small
\setlength{\tabcolsep}{6pt}
\begin{tabular}{@{}l c c c@{}}
\toprule
\textbf{Disposition} & \textbf{Acuity} & \textbf{Levine 4 level} & \textbf{Gaber ESI} \\
\midrule
EMS 911 now                 & 1 & Emergent  & 1 \\
ED now                      & 2 & Emergent  & 2 \\
ED/UCC now (PCP triage)     & 2 & 1 day     & 2 \\
HCP ${\le}4$\,h (PCP triage) & 3 & 1 day    & 3 \\
PCP now                     & 3 & 1 day     & 3 \\
See PCP ${\le}24$\,h        & 3 & 1 day     & 3 \\
Call PCP ${\le}24$\,h       & 3 & 1 day     & 3 \\
See PCP ${\le}3$\,d         & 4 & 1 week    & 4 \\
See PCP ${\le}2$\,wk        & 4 & 1 week    & 4 \\
Home care                   & 5 & Self-care & 5 \\
\bottomrule
\end{tabular}
\end{table}

\FloatBarrier
\section{Prior methods and external vignette sets}\label{app:reproduction}

\begin{figure}[!htbp]
  \centering
  \includegraphics[width=0.7\linewidth]{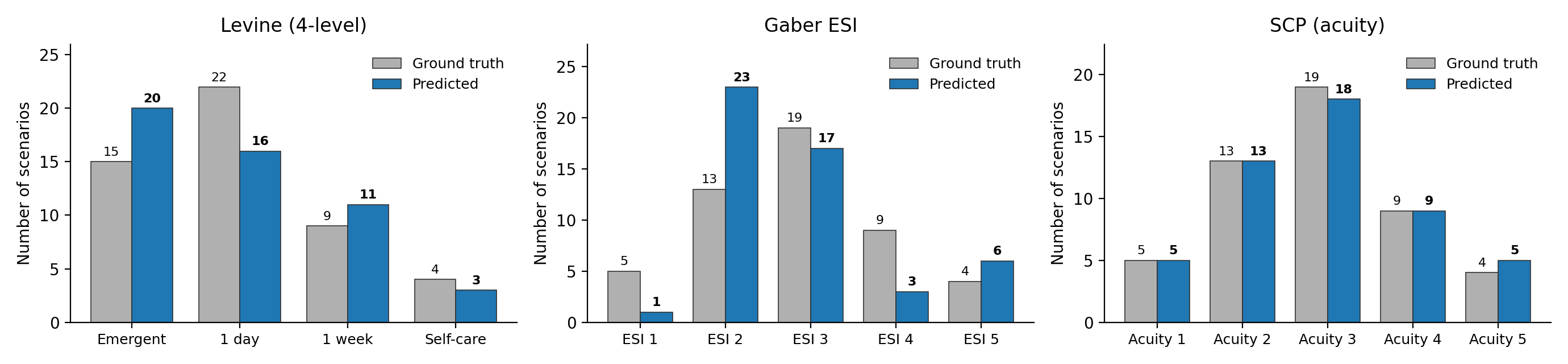}
  \caption{Ground-truth vs.\ predicted label distributions for the two reproduced methods and ALLMTS
  (\texttt{gpt-5.2}, 50 scenarios). Both compress predictions toward the middle of their scale: Gaber
  under-uses the extreme ESI~1 and ESI~4 bins while over-populating ESI~2, and Levine over-triages into
  \emph{Emergent}.}
  \label{fig:reproduction-bias}
\end{figure}

Figure~\ref{fig:reproduction-bias} shows the label distributions behind
Table~\ref{tab:reproduction-metrics}, and Table~\ref{tab:allmts-external-vignettes}
reports ALLMTS run directly on the external vignette sets
(Section~\ref{sec:results}).

\begin{table}[!htbp]
\caption{ALLMTS (\texttt{gpt-5.2}, low reasoning) run directly on the non-pediatric subset of external vignette sets, scored via-acuity in each set's native label space.}
\label{tab:allmts-external-vignettes}
\centering
\small
\setlength{\tabcolsep}{6pt}
\begin{tabular}{@{}l c c c c c c@{}}
\toprule
\textbf{Dataset} & \textbf{N} & \textbf{Acc} & \textbf{F1} & \textbf{Under} & \textbf{Over} & \textbf{Disc} \\
\midrule
Levine et al.    & 43 & 0.41 & 0.46 & 0.34 & 0.25 & 0.59 \\
Semigran et al.  & 37 & 0.58 & 0.64 & 0.29 & 0.13 & 0.42 \\
\bottomrule
\end{tabular}
\end{table}

\FloatBarrier
\section{Label distribution}\label{app:label-distribution}

\begin{figure}[!htbp]
\centering
\begin{subfigure}[t]{0.49\linewidth}
  \includegraphics[width=\linewidth]{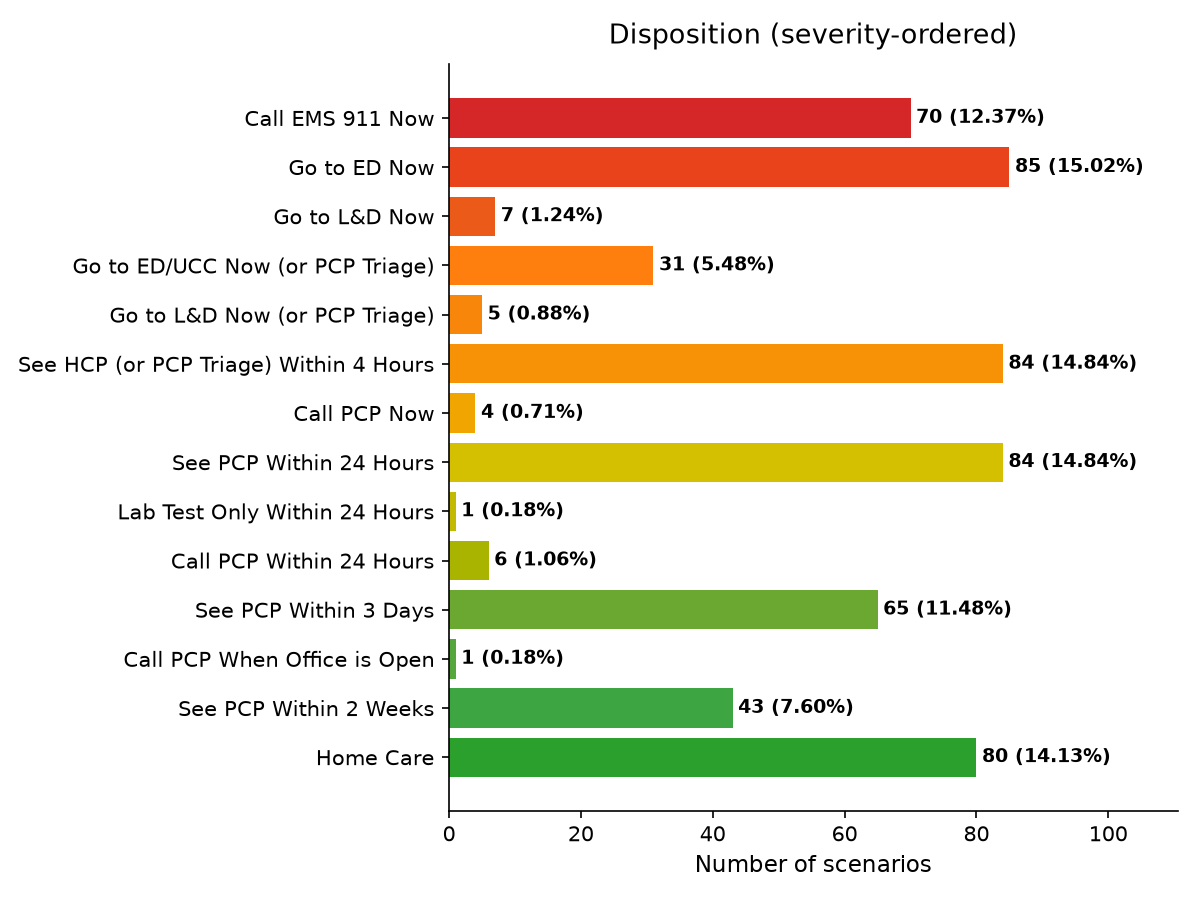}
  \caption{Disposition (14 classes, severity-ordered)}
\end{subfigure}\hfill
\begin{subfigure}[t]{0.49\linewidth}
  \includegraphics[width=\linewidth]{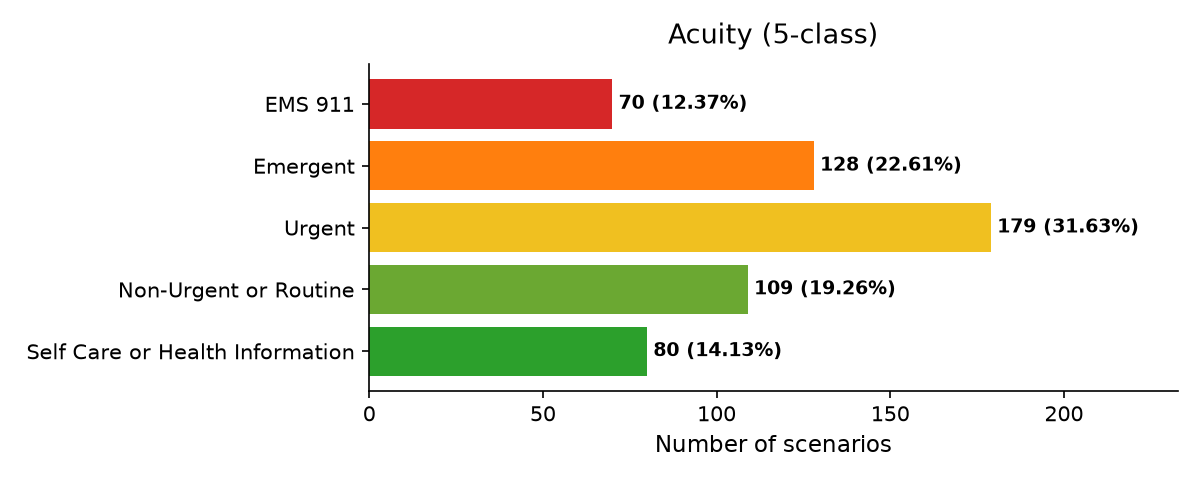}
  \caption{Acuity (5-class)}
\end{subfigure}

\medskip
\begin{subfigure}[t]{0.49\linewidth}
  \includegraphics[width=\linewidth]{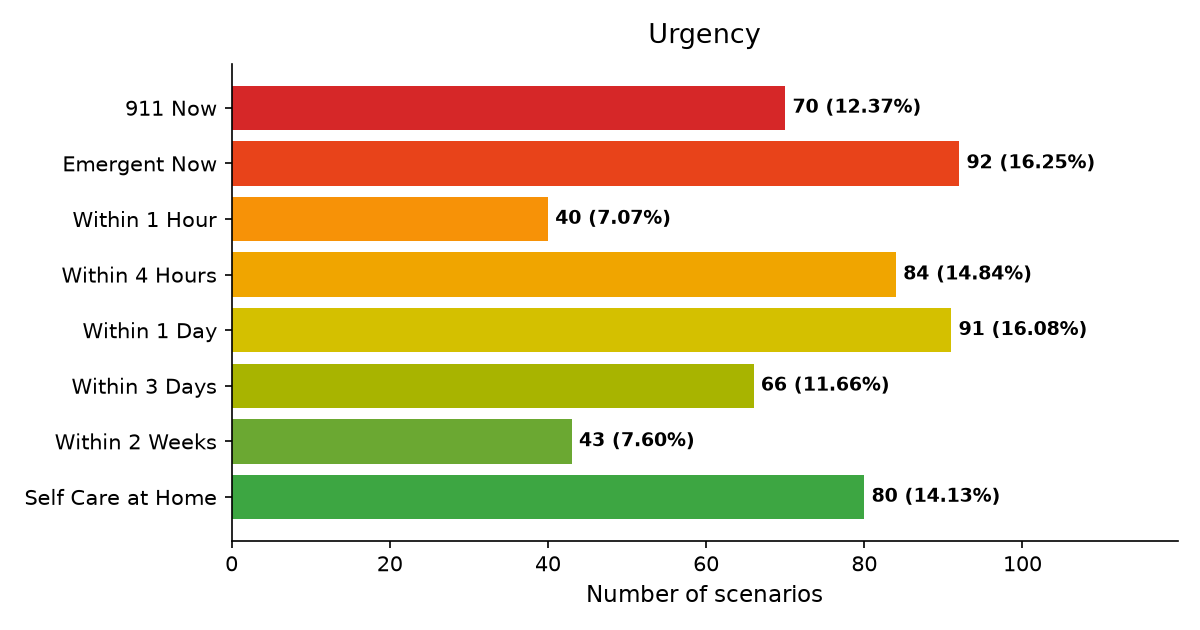}
  \caption{Urgency}
\end{subfigure}\hfill
\begin{subfigure}[t]{0.49\linewidth}
  \includegraphics[width=\linewidth]{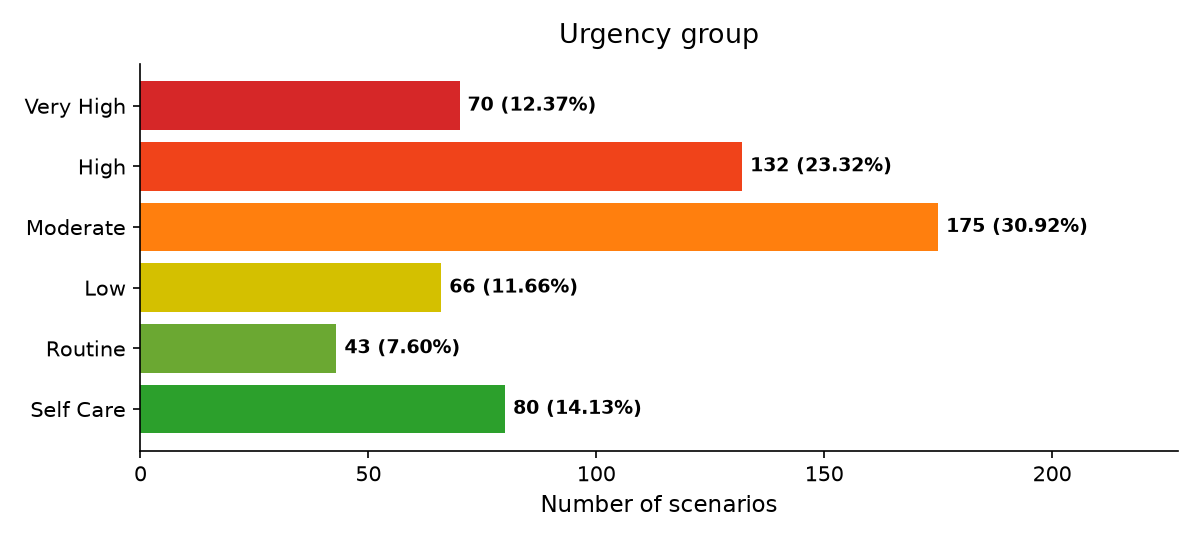}
  \caption{Urgency group}
\end{subfigure}
\caption{Reference label distribution of the 566-scenario dataset across the four label views, severity ordered (most urgent first) and coloured as in the confusion matrices. The distribution is head heavy with a long thin tail in the fine-grained disposition space; coarsening collapses the tail into populated bands.}
\label{fig:dataset-dist}
\end{figure}

\FloatBarrier
\section{Run-to-run stability}\label{app:stability}

\begin{figure}[!htbp]
\centering
\includegraphics[width=\linewidth]{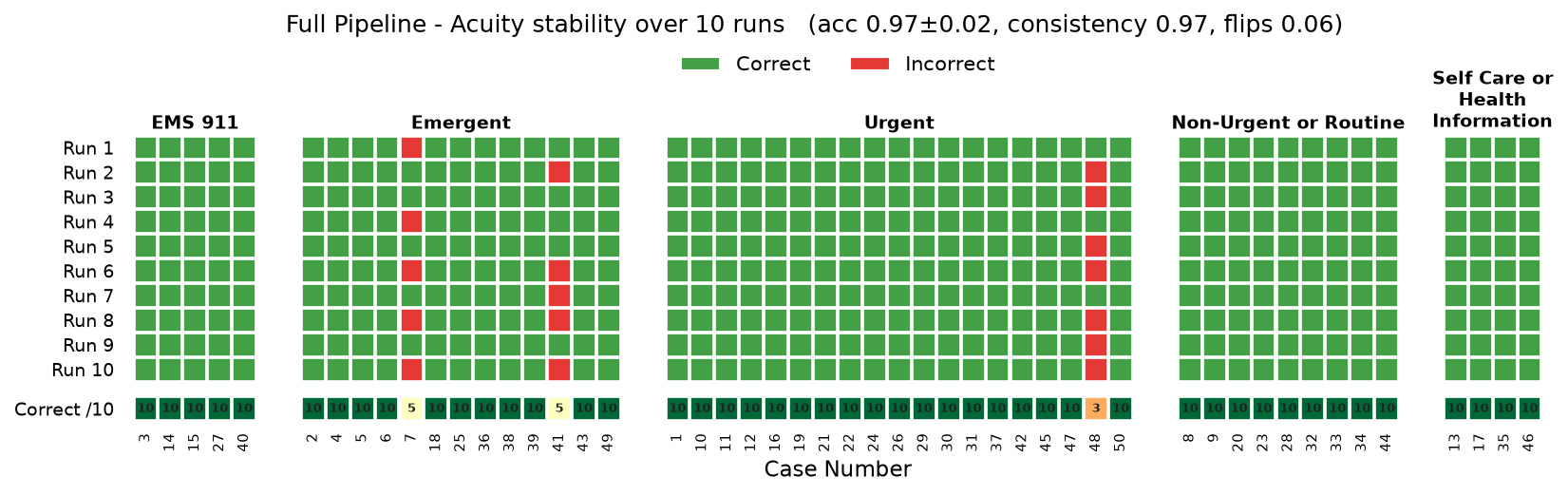}
\caption{%
  Run-to-run stability of ALLMTS. Nearly every cell is green: ALLMTS reproduces the reference
  acuity band on all ten runs for 47 of 50 scenarios, and no scenario is wrong on every run.
}
\label{fig:stability-scp}
\end{figure}

\begin{figure}[!htbp]
\centering
\includegraphics[width=\linewidth]{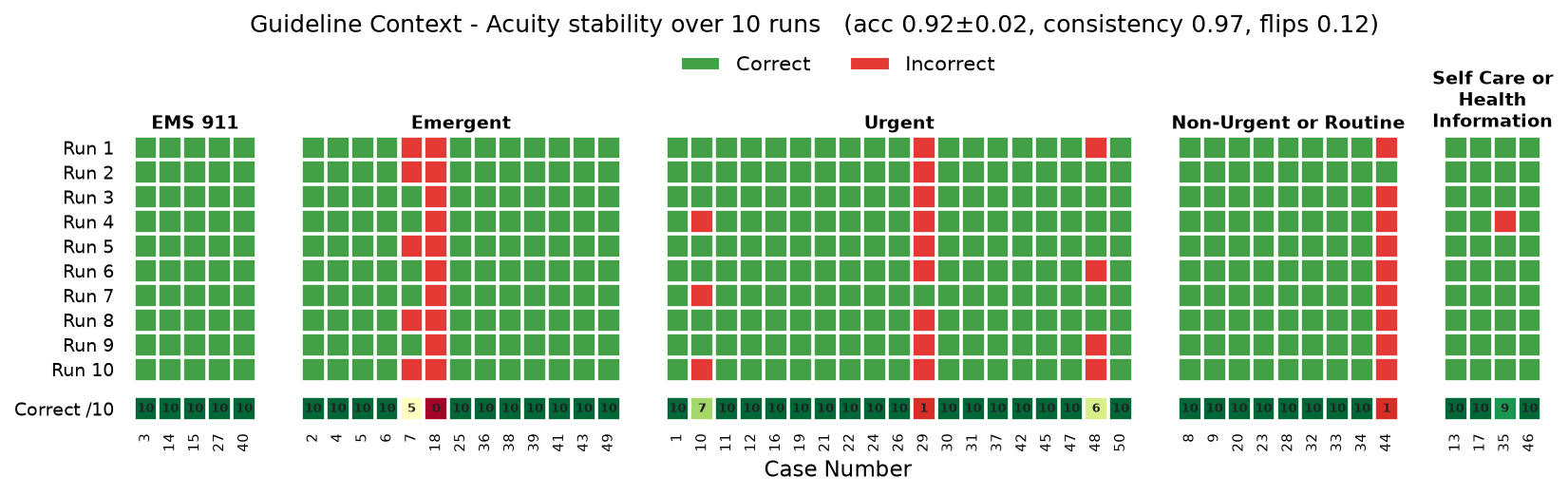}
\caption{%
  Run-to-run stability of the guideline-context ablation. Reading the selected guideline
  holistically is nearly as consistent as executing it ($0.970$ versus $0.974$ for
  ALLMTS) but less accurate: $43$ of $50$ scenarios are correct on every run and one is
  wrong on all ten.
}
\label{fig:stability-gc}
\end{figure}

\begin{figure}[!htbp]
\centering
\includegraphics[width=\linewidth]{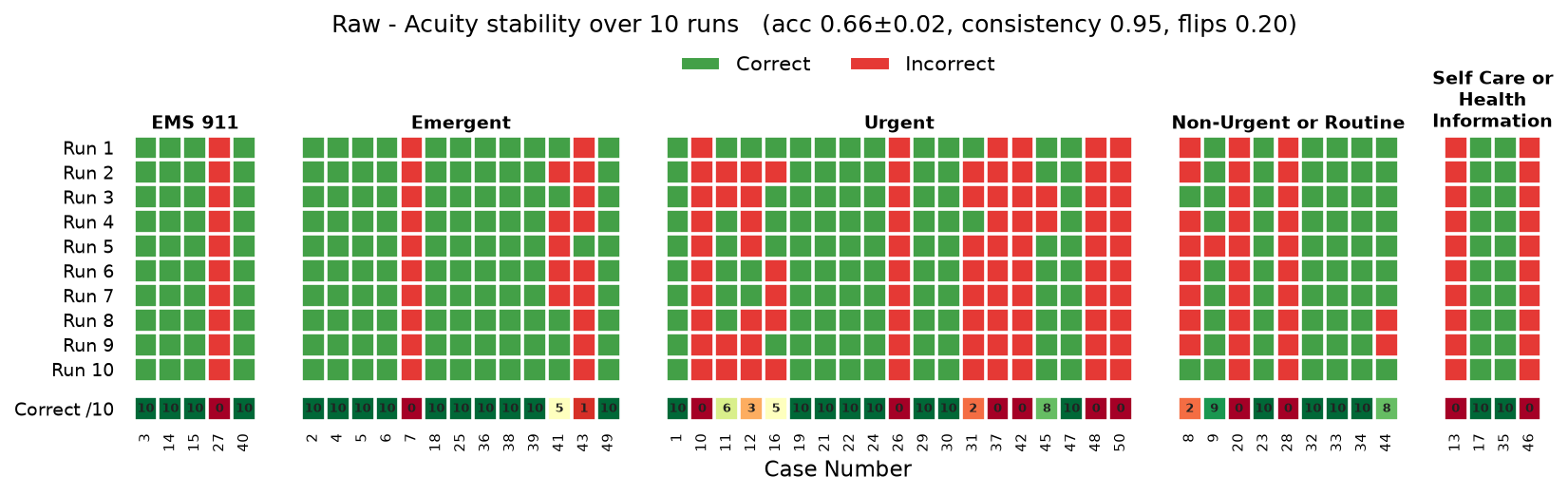}
\caption{%
  Run-to-run stability of the Raw LLM, shown for comparison with ALLMTS (Figure~\ref{fig:stability-scp}) and Web Search (Figure~\ref{fig:stability-ws}).
}
\label{fig:stability-raw}
\end{figure}

\begin{figure}[!htbp]
\centering
\includegraphics[width=\linewidth]{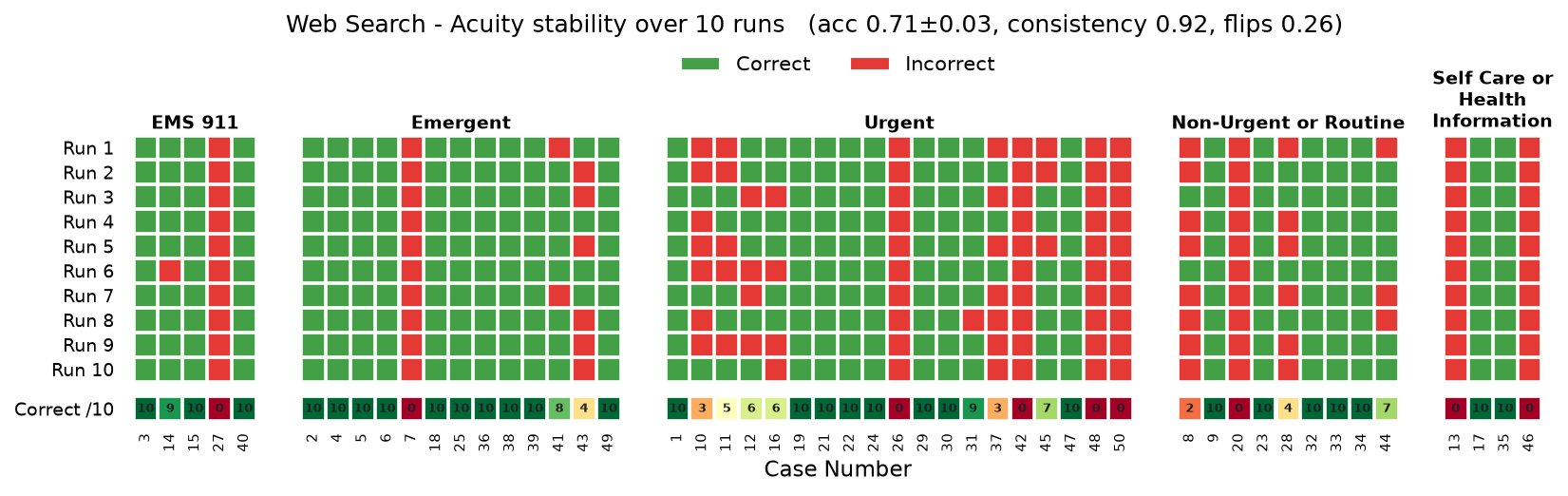}
\caption{%
    Run-to-run stability of the Web Search baseline. Web Search is the least consistent of the four methods (flip rate 0.26), with 13 of 50 scenarios changing prediction across runs.
}
\label{fig:stability-ws}
\end{figure}

Figures~\ref{fig:stability-scp}--\ref{fig:stability-ws} give the stability grids described in
Section~\ref{sec:results}, one per method. Cells are red where the run missed the reference
acuity band, columns are grouped by reference acuity (most urgent first) and ordered by
scenario index, and each title reports mean accuracy $\pm$ its standard deviation across
runs, modal prediction consistency, and the flip rate (the fraction of scenarios whose
prediction changed across runs).

\FloatBarrier
\section{Confusion matrices}\label{app:confusion-matrices}

\begin{figure}[!htbp]
\centering
\begin{subfigure}[t]{0.48\linewidth}
  \includegraphics[width=\linewidth]{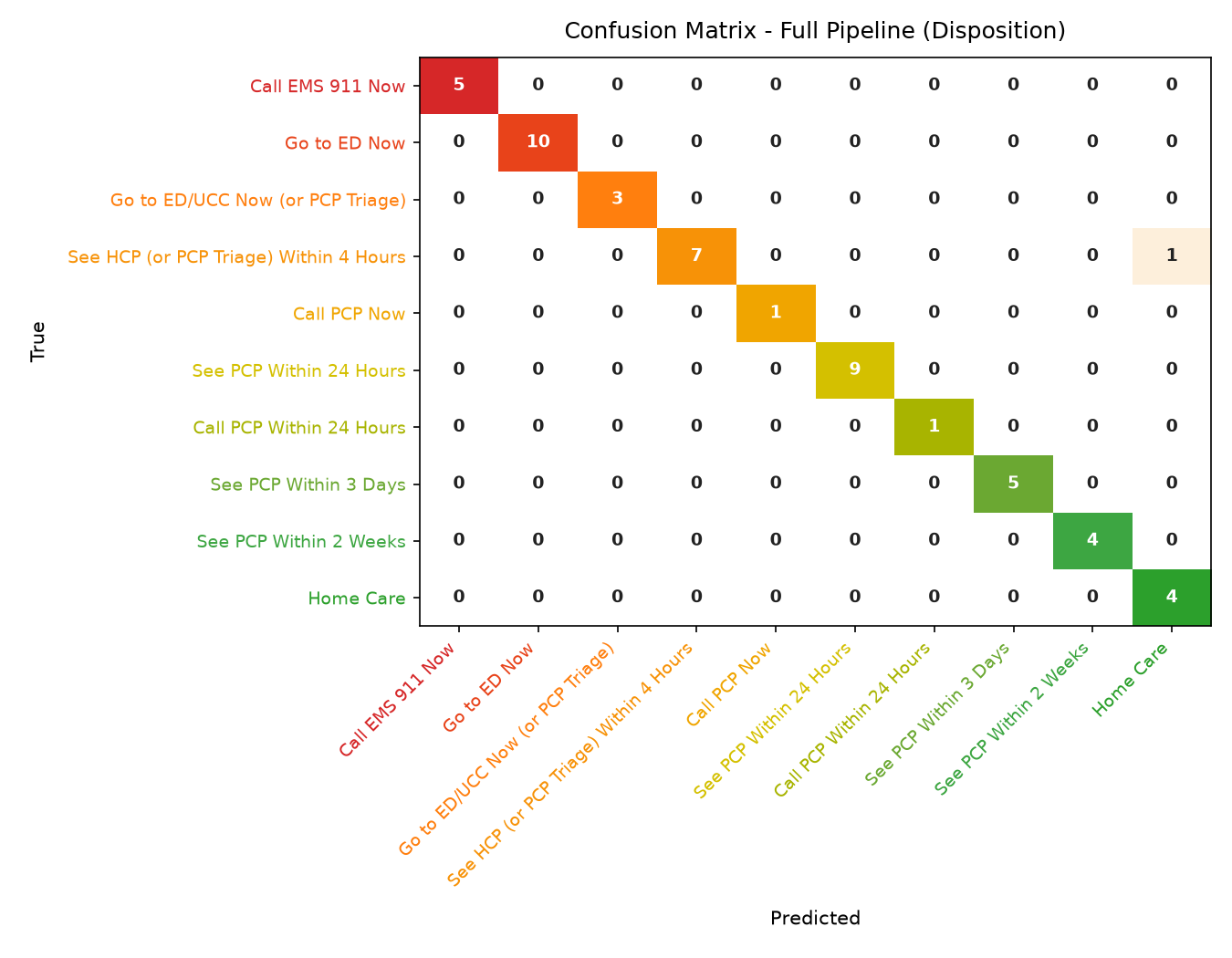}
  \caption{ALLMTS, Disposition (acc.\ 0.98)}
\end{subfigure}\hfill
\begin{subfigure}[t]{0.48\linewidth}
  \includegraphics[width=\linewidth]{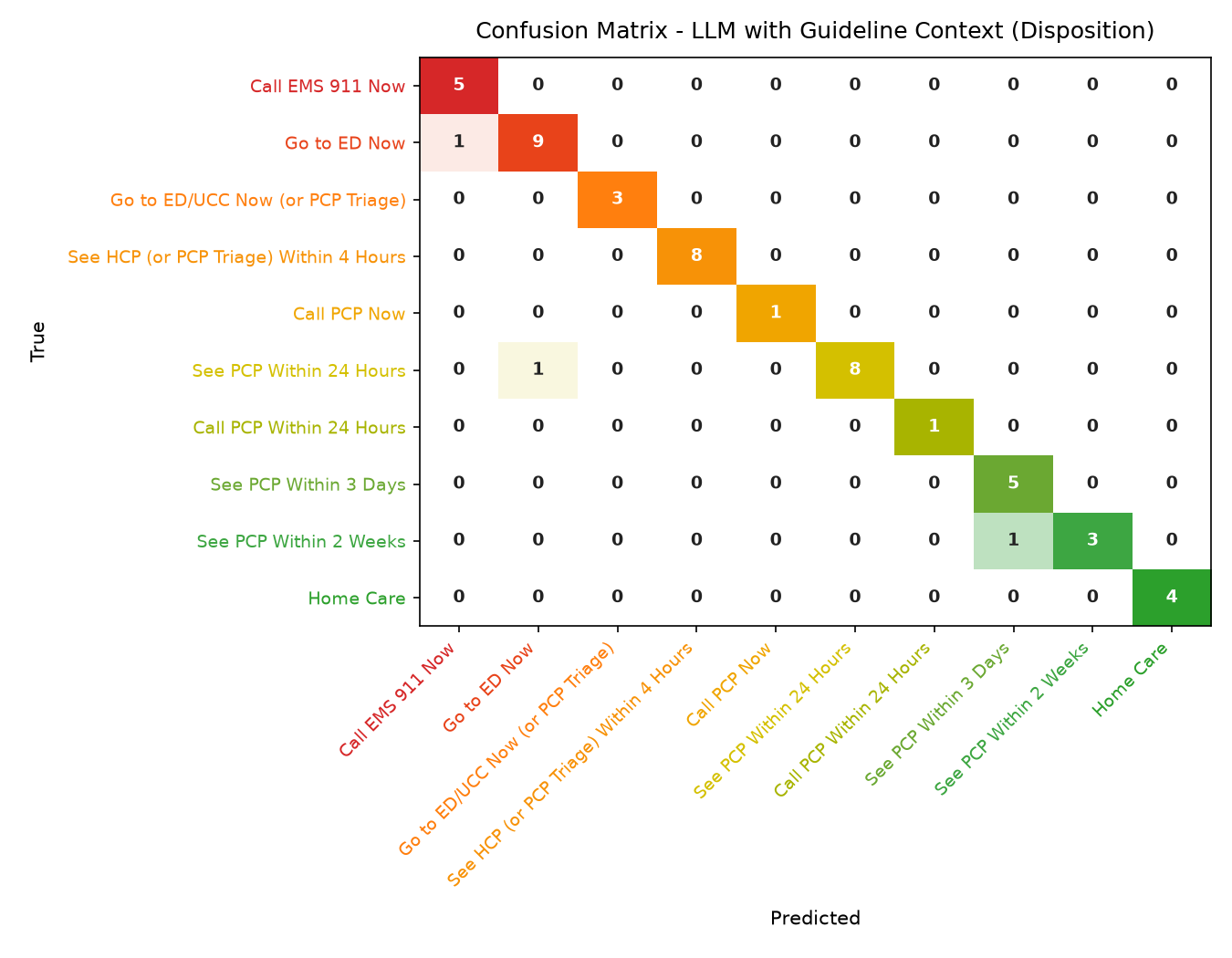}
  \caption{Guideline Ctx, Disposition (acc.\ 0.94)}
\end{subfigure}

\medskip
\begin{subfigure}[t]{0.48\linewidth}
  \includegraphics[width=\linewidth]{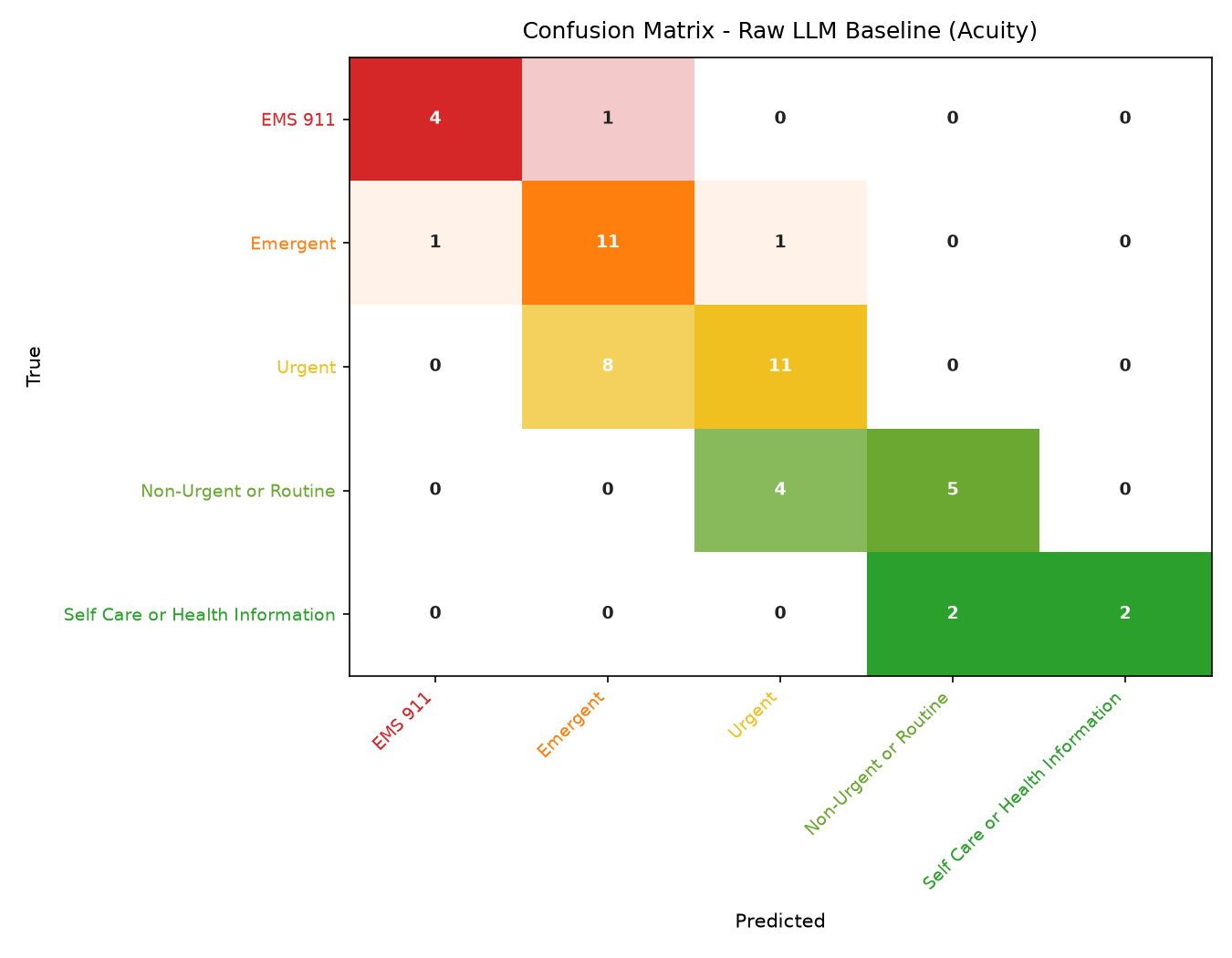}
  \caption{Raw LLM, Acuity (acc.\ 0.66)}
\end{subfigure}\hfill
\begin{subfigure}[t]{0.48\linewidth}
  \includegraphics[width=\linewidth]{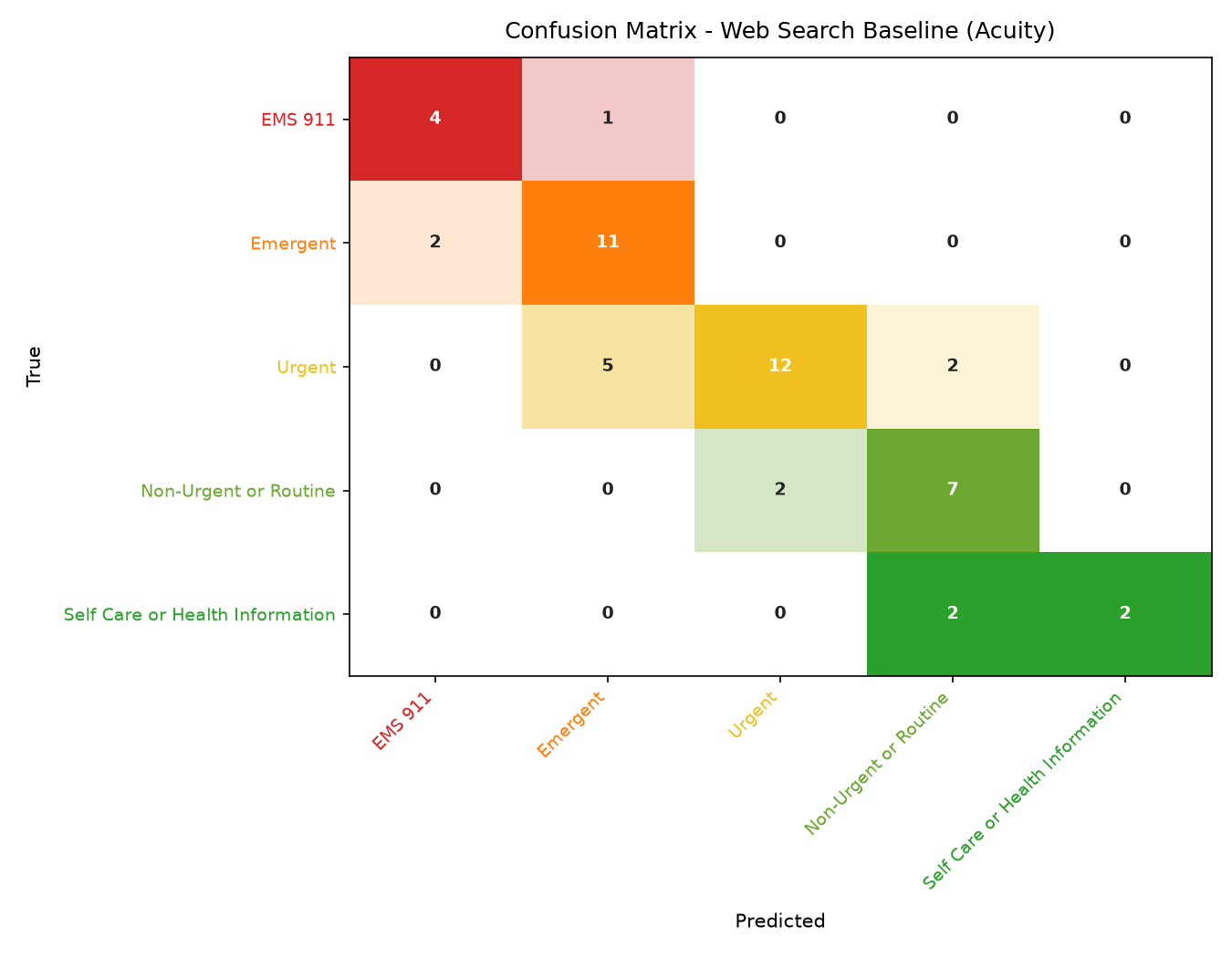}
  \caption{Web Search, Acuity (acc.\ 0.72)}
\end{subfigure}
\caption{Confusion matrices on the 50-scenario set. The guideline grounded methods stay tightly concentrated along the diagonal, whereas both baselines exhibit an upward spread toward more urgent predictions, indicating a tendency to over-triage.}
\label{fig:cm}
\end{figure}

\begin{figure}[!htbp]
\centering
\begin{subfigure}[t]{0.48\linewidth}
  \includegraphics[width=\linewidth]{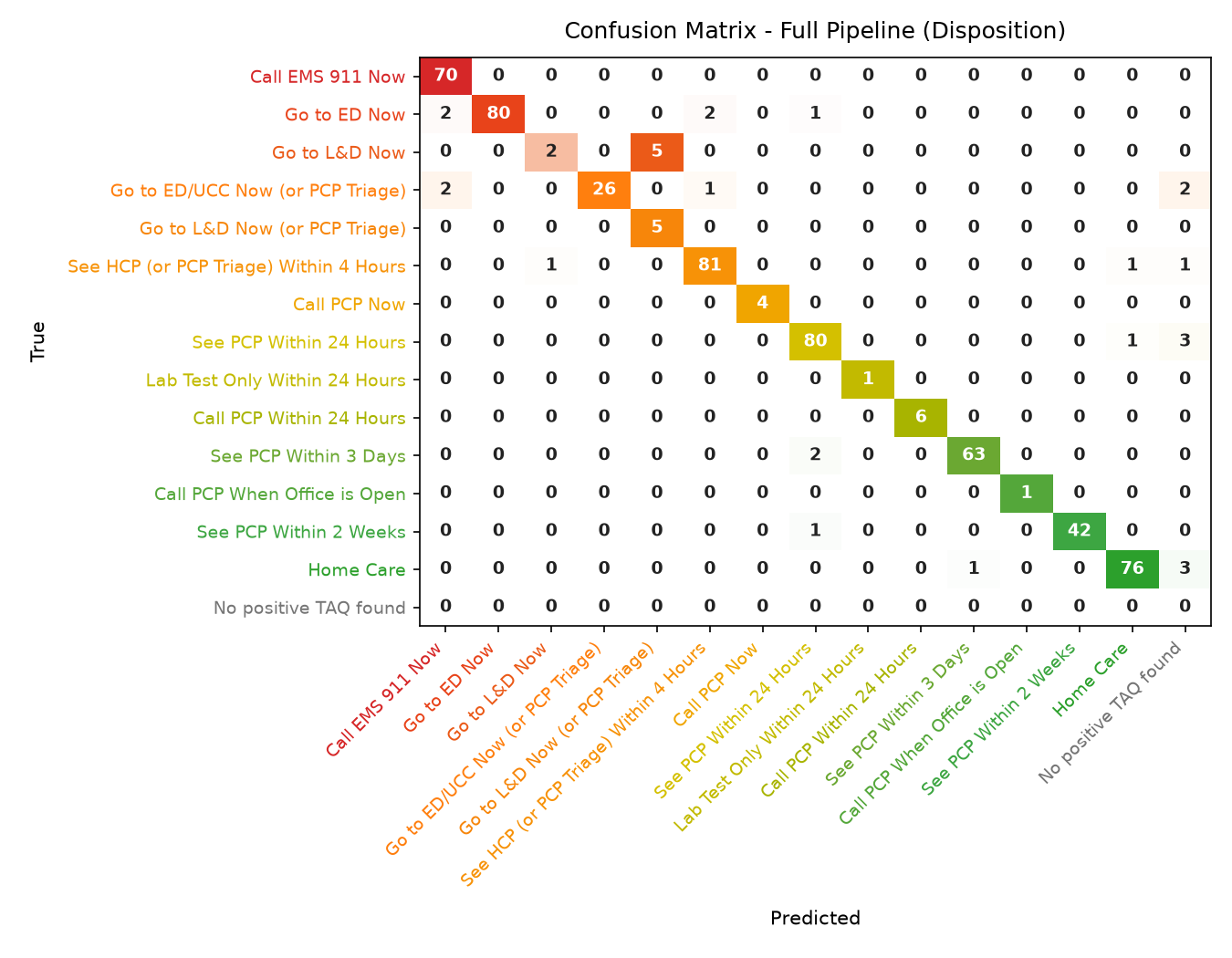}
  \caption{ALLMTS, Disposition (acc.\ 0.95)}
\end{subfigure}\hfill
\begin{subfigure}[t]{0.48\linewidth}
  \includegraphics[width=\linewidth]{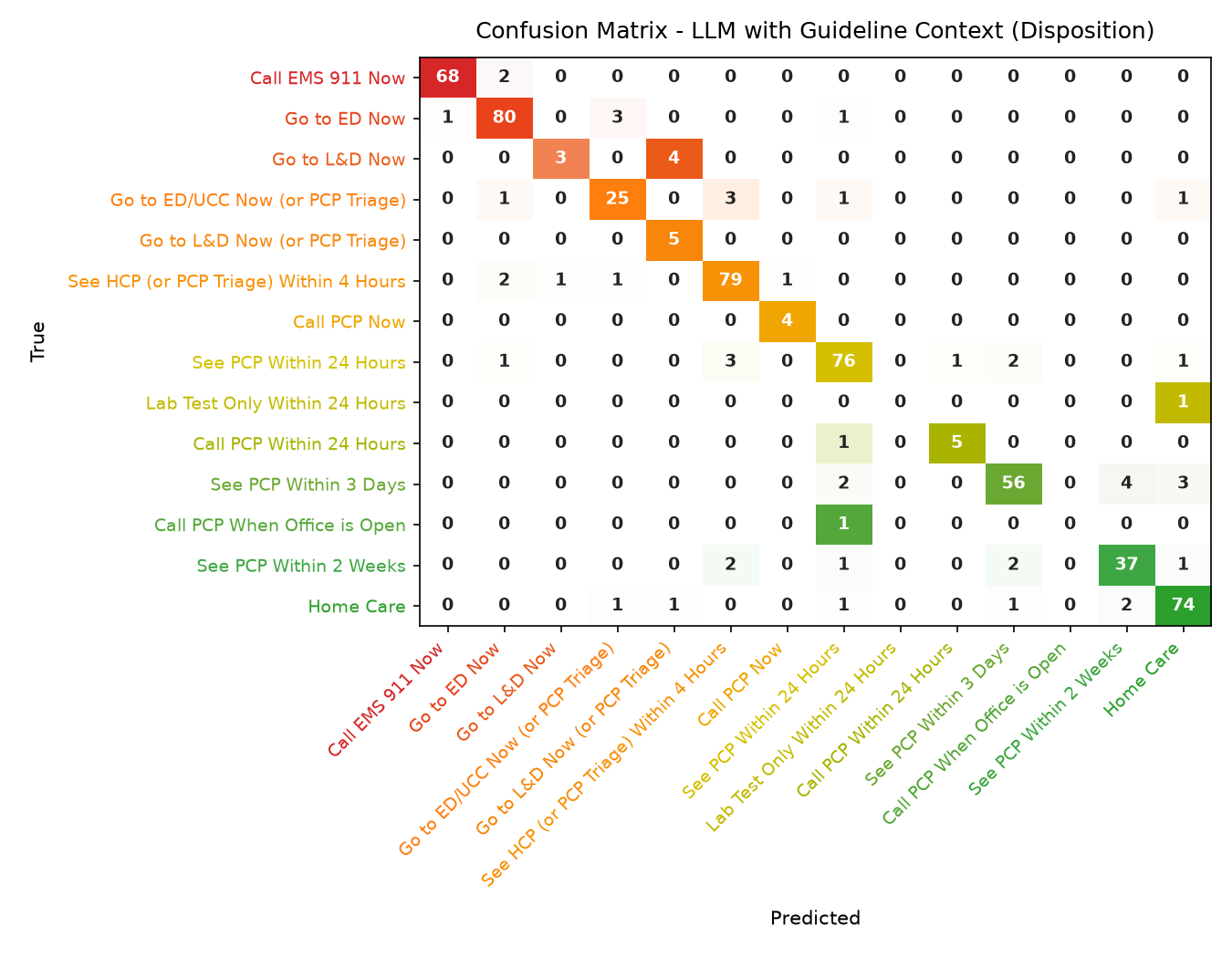}
  \caption{Guideline Ctx, Disposition (acc.\ 0.90)}
\end{subfigure}

\medskip
\begin{subfigure}[t]{0.48\linewidth}
  \includegraphics[width=\linewidth]{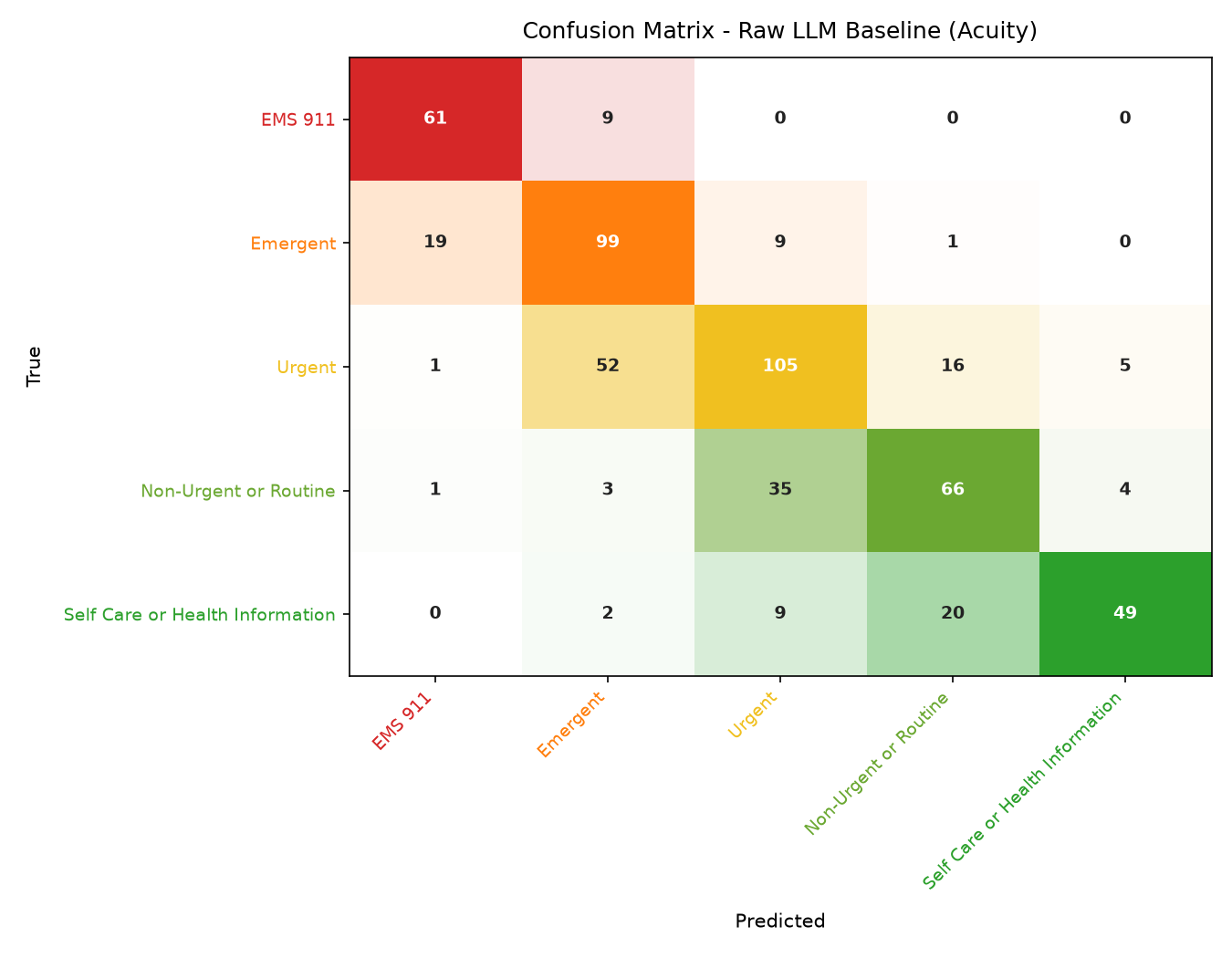}
  \caption{Raw LLM, Acuity (acc.\ 0.67)}
\end{subfigure}\hfill
\begin{subfigure}[t]{0.48\linewidth}
  \includegraphics[width=\linewidth]{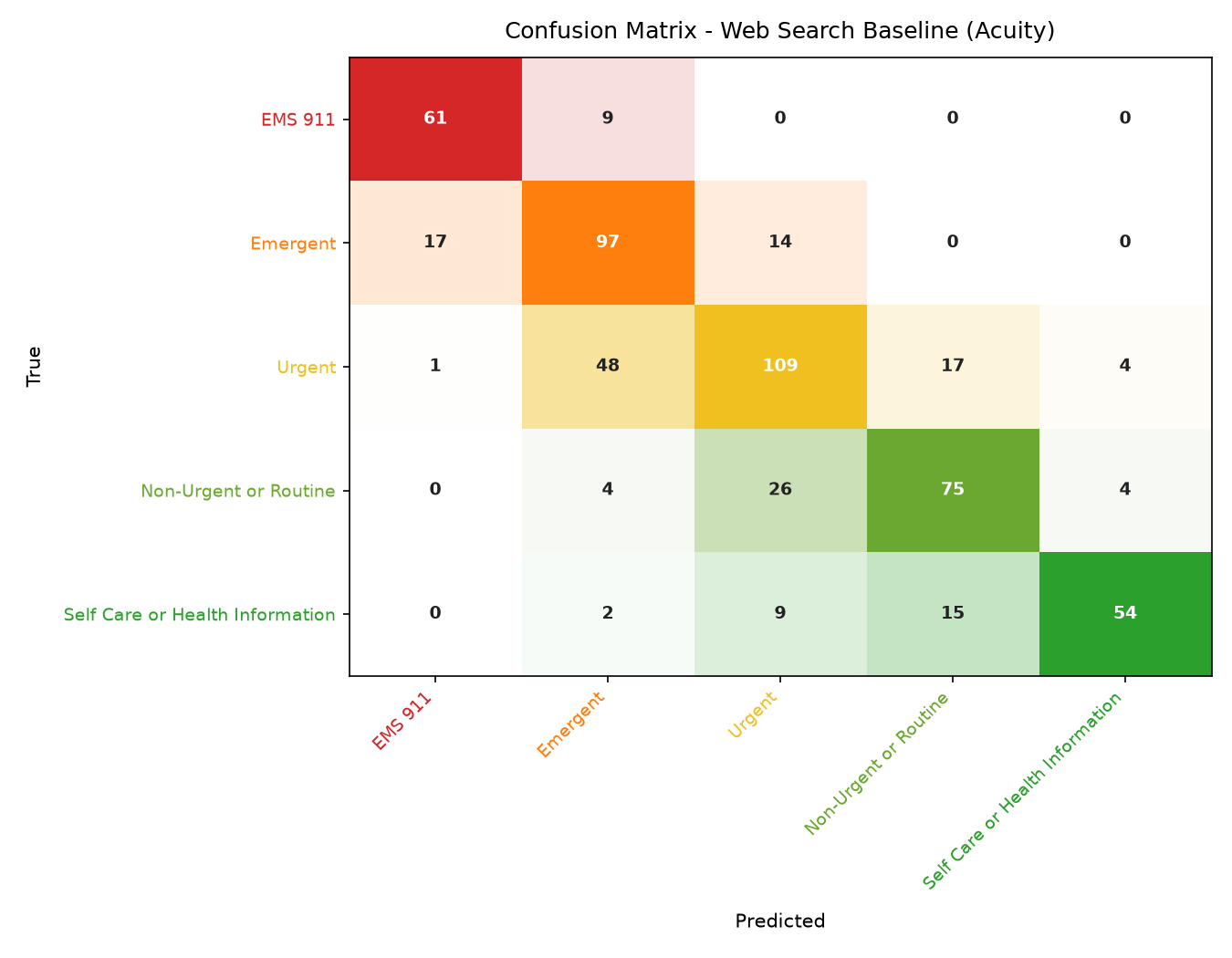}
  \caption{Web Search, Acuity (acc.\ 0.70)}
\end{subfigure}
\caption{Confusion matrices on the full 566-scenario dataset. ALLMTS's mass stays on the diagonal across the wider 14 disposition vocabulary; the trailing \emph{No positive TAQ found} class collects cases where the pipeline found no positive TAQ. Both baselines again spread upward into more urgent predictions.}
\label{fig:cm-566}
\end{figure}

Figures~\ref{fig:cm} and~\ref{fig:cm-566} give the ad-hoc confusion matrices for both
evaluation sets, each method shown on its strongest view: ALLMTS and the
guideline-context ablation on the disposition space, the two baselines on the acuity
view.

\end{document}